\documentclass[runningheads]{llncs}
\usepackage{graphicx}
\usepackage{subfigure}
\usepackage{diagbox}
\usepackage{longtable}
\usepackage{algorithm,algorithmicx,algpseudocode}
\usepackage{booktabs}
\usepackage{amsmath,amssymb,amsfonts}
\usepackage{xcolor}
\usepackage{multirow}

\renewcommand{\baselinestretch}{0.99} \normalsize

\begin{document}
%
\title{Data Augmentation for Graph Convolutional Network on Semi-Supervised Classification}
%
%

\author{Zhengzheng Tang$^{\dag}$\inst{,1,2} \and
Ziyue Qiao$^{\dag}$\inst{,1,2} \and
Xuehai Hong$^{\ddag}$\inst{3,1} \and 
Yang Wang\inst{2} \and
Fayaz Ali Dharejo\inst{1,2} \and
Yuanchun Zhou\inst{2} \and
Yi Du\inst{2}
}
\renewcommand{\thefootnote}{\fnsymbol{footnote}}
\footnotetext{$^\dag$Zhengzheng Tang and Ziyue Qiao contributed equally to this work.}
\footnotetext{$^{\ddag}$Xuehai Hong is the Corresponding author.}
\authorrunning{Z. Tang et al.}
%
\institute{Computer Network Information Center, Chinese Academy of Sciences, Beijing \and
University of Chinese Academy of Sciences, Beijing\and
Institute of Computing Technology, Chinese Academy of Sciences, Beijing
\email{\{tangzhengzheng,qiaoziyue,wangyang\}@cnic.cn}, \email{hxh@ict.ac.cn},  \email{\{fayazdharejo,zyc,duyi\}@cnic.cn}}


%
\maketitle              
\begin{abstract}

Data augmentation aims to generate new and synthetic features from the original data, which can identify a better representation of data and  improve the performance and generalizability of downstream tasks.
However, data augmentation for graph-based models remains a challenging problem, as graph data is more complex than traditional data, which consists of two features with different properties: graph topology and node attributes.
In this paper, we study the problem of graph data augmentation for Graph Convolutional Network (GCN) in the context of improving the node embeddings for semi-supervised node classification.
Specifically, we conduct cosine similarity based cross operation on the original features to create new graph features, including new node attributes and new graph topologies, and we combine them as new pairwise inputs for specific GCNs.  
Then, we propose an attentional integrating model to weighted sum the hidden node embeddings encoded by these GCNs into the final node embeddings. We also conduct a disparity constraint on these hidden node embeddings when training to ensure that non-redundant information is captured from different features.
Experimental results on five real-world datasets show that our method  improves the classification accuracy with a clear margin(+2.5\% - +84.2\%) than the original GCN model.


\end{abstract}
\section{Introduction}

Data augmentation can create several new feature spaces and increase the amount of training data without additional ground truth labels, which has been widely used to improve the performance and generalizability of downstream predictive models. Many works have proposed data augmentation technologies on different types of features, such as images\cite{8363576,8242527,moreno2018forward}, texts\cite{8563226,8935169}, vectorized features\cite{song2018novel,fawaz2018data}, etc. However, how to effectively augment graph data remain a challenging problem, as graph data is more complex and has non-Euclidean structures. Graph Neural Network(GNN) is a family of graph representation learning approaches that encode node features into low-dimensional representation vectors by aggregating local neighbors' information, it has drawn increasing attention in recent years, due to the superior performance on graph data mining \cite{kipf2016semi,wang2019heterogeneous,Wang2020AM}.


For graph-based semi-supervised classification, the goal is to use the given graph data to predict the labels of unlabeled nodes. The given graph data usually consists of graph topology, node attributes(also called node features in some literature, we use node attributes to avoid the confusion with graph feature), as well as the labels of a subset node.
Despite the labels, graph data can be specifically described as two graph features: an adjacency matrix of graph topology $A\in \mathbb{R}^{N \times N}$ and a node attribute matrix $X\in \mathbb{R}^{N \times d}$, where $N$ is the total number of nodes, and $d$ is the dimension of node attribute. 
GNN models conduct on both of these two features simultaneously and fuse them into the final node embedding by stacking several aggregation layers. The whole model can be formulated as a multi-layer graph encoder $Z=G(A,X)$, where $Z\in \mathbb{R}^{N \times h}$ is the output node embedding matrix and $h$ is the dimension of node embedding. 
In this work, we consider the most popular and representative GNN: Graph Convolutional Network(GCN), proposed by Kipf et al. \cite{kipf2016semi}, which is the state-of-the-art model for semi-supervised node classification.
It uses an efficient layer-wise propagation rule based on a first-order approximation of spectral convolutions on graphs.
The encoder function $Z =G(A,X)$ of a $L$-layers' GCN can be specified as:

\begin{equation}
    Z=G(A,X)= \sigma(\hat{A}...\sigma(\hat{A}\sigma(\hat{A}XW^{(0)})W^{(1)})...W^{(L)})
\end{equation}

where $L$ is the number of layers. $W^{(i)}$ is the weight matrix of the $i$-th layer of GCN, $\sigma$ denotes an activation function. $\hat{A}= \tilde{D}^{-\frac{1}{2}}\tilde{A}\tilde{D}^{-\frac{1}{2}}$, $\tilde{A} = A+I_N$, $I_N$ is the identity matrix and $\hat{D}$ is the diagonal degree matrix of $\tilde{A}$.

However, a fact is that as the pairwise input for the GCN model, both the original features $A$ and $X$ may not be positive correlated with the node labels, while GCN can not adequately learn the importance of these two features to extract the most correlated information, which dampens the performance of GCN on the classification task.
Data augmentation can create new feature spaces and preserve the information in original graph data in multiple facets, some of which may contribute useful information to node classification.
This leads to the question: besides the original graph features $A$ and $X$, can we create new pairs of adjacency matrices and attribute matrices and adaptively choose some effective ones as new feature inputs for GCN models?

Many prior studies\cite{fawaz2018data,8659168} in data augmentation are to capture the interactions between features by taking addition, subtraction, or cross product of two original features, which are suitable for tensorial features.
The major obstacle in graph data is that the original features, graph topology, and node attributes, are two types of data, one is usually encoded by position in Euclidean space, while the other is encoded by node connectivity in non-Euclidean space. It is difficult to take combination operations on these two features to create new features. 
Some work\cite{rong2019dropedge,JMLR:v15:srivastava14a,wang2020nodeaug} proposes different strategy of adding or removing edges to improve the robustness of GCN. However, these augmentation methods are limited to modifying just a part of the node featuring in the graph, which is unable to create a brand new feature space of the whole graphs for GCN.

In this paper, we first create multiple new graph  topologies and node attributes from the given graph data and propose different combinations of them as inputs for specific GCN models. Then, the output node embeddings of different GCN models are assigned with different weights via an attention mechanism, to sum up to the final node embeddings. In the training, an independence measurement-based disparity constraint is integrated into the objective function to capture diverse information from different features. In this way, extensive information from the original graph is encoded into the final node embeddings to improve the semi-supervised node classification task. The main contributions of our work are summarized as follows:

\begin{enumerate}
\item We propose a graph data augmentation strategy to create new pairwise graph inputs for the GCN model by designing new node attributes and graph topologies from the original graph features.
\item We propose an attentional integrating model, which can learn the importance of different hidden node embeddings encoded from various pairwise graph inputs via specific GCNs, and integrate them into the final node embeddings.
\item We propose a Hilbert-Schmidt independence criterion-based disparity constraint to increase the independence between the node embeddings encoded from various pairwise graph inputs and capture more diverse information.
\item We conduct experiments to evaluate the performance of our proposed method on five datasets. 
Our improvement over original GCN is +2.5\% - +84.2\%.
\end{enumerate}

\section{Proposed Method}
In this section, we introduce the graph data augmentation strategies for GCN, then we investigate the availability of our augmented features by intuitive cases. Finally, we introduce the whole model including the attentional integrating model and the disparity constraint.

\subsection{Data Augmentation Strategy}
\label{sec:DA}
Given the original features $A$ and $X$ of graph data, we aim to reconstruct the whole graph topology and node attributes. A naive and widely used way of data augmentation operation is cross operation, we first conduct cosine similarity-based cross operation on $A$ and $X$ to create two new features, which carry the information of global proximity of nodes with others in the views of local topology and node attributes. Specifically, 
for each row in $A$ and $X$, we calculate the cosine similarities of it with all the other rows and concatenate these similarities as new features of its corresponding node. Finally, the new features matrices $A_C$ and $X_C$ of the graph can be formulated as: 

\begin{equation}
    A_{C_{ij}} = \frac{A_i\cdot A_j}{\|A_i\| \|A_j\|}, \quad X_{C_{ij}} = \frac{X_i\cdot X_j}{\|X_i\| \|X_j\|}.
\end{equation}


where  $A_C\in \mathbb{R}^{N \times N}$, $X_C\in \mathbb{R}^{N \times N}$, $A_{C_{ij}}$ and $X_{C_{ij}}$ is the element in the $i$-th row and $j$-th column of $A_C$ and $X_C$ respectively, $A_i$ and $X_i$ is the $i$-th row of $A$ and $X$ respectively. We consider $A_C$ and $X_C$ as new node attribute matrices, as for each node, its corresponding row in $A_C$ preserves the information of global structural proximity with other nodes, and that in $X_C$ preserves the information of global proximity of attribute with other nodes. To some extent, these information can be regarded as different types of node attributes. 

Further, we use the obtained $A_C$ and $X_C$ to construct $k$-nearest neighbor graphs  $A_T \in \{0,1\}^{N\times N}, X_T \in \{0,1\}^{N\times N}$, that is, we set the largest $k$ elements in each row as 1 and set other elements as 0. $A_T$ and $X_T$ are considered as new adjacency matrices, where each edge in $A_T$ represents the connecting nodes are similar in local topology and each edge in $X_T$ represents the connecting nodes are similar in node attribute. 

Finally, we combine these attribute features and adjacency features to create 9 different inputs for GNN model, as shown in the Table \ref{combination}:

\begin{table*}
\centering
\caption{Different combinations of six graph features $A, X, A_C, X_C, A_T, X_T$ as inputs for GNN model. Adj. means the adjacency matrices, Att. means the attribute matrices. $G_i(\cdot, \cdot)$ represent the specific GNN encoder for the $i$-th combination of features.}
\begin{tabular}{c|c|c|c}
\hline
\diagbox{Adj.}{Att.} & $X$       & $A_C$       & $X_C$       \\ \hline
$A$     & $G_1(A, X)$ & $G_2(A,A_C)$ & $G_3(A, X_C)$ \\ \hline
$A_T$   & $G_4(A_T, X)$ & $G_5(A_T, A_C)$ & $G_6(A_T, X_C)$  \\ \hline
$X_T$   & $G_7(X_T,X)$  & $G_8(X_T, A_C)$  & $G_9(X_T, X_C)$  \\ \hline

\end{tabular}
\label{combination}
\end{table*}

Noted that the adjacency matrix is usually very sparse, making the cosine similarity matrix sparse, too. So before the process of data augmentation, we first use the update rule proposed in \cite{2017Fast} through the original adjacency matrix $A$ to build new edges between neighbors within 2-hop links, and upgrade $A$ as a denser high-order adjacency matrix.


\subsection{Feature Availability Investigation}
\begin{figure*}
\centering
\subfigure[X]{\includegraphics[width=3.5cm]{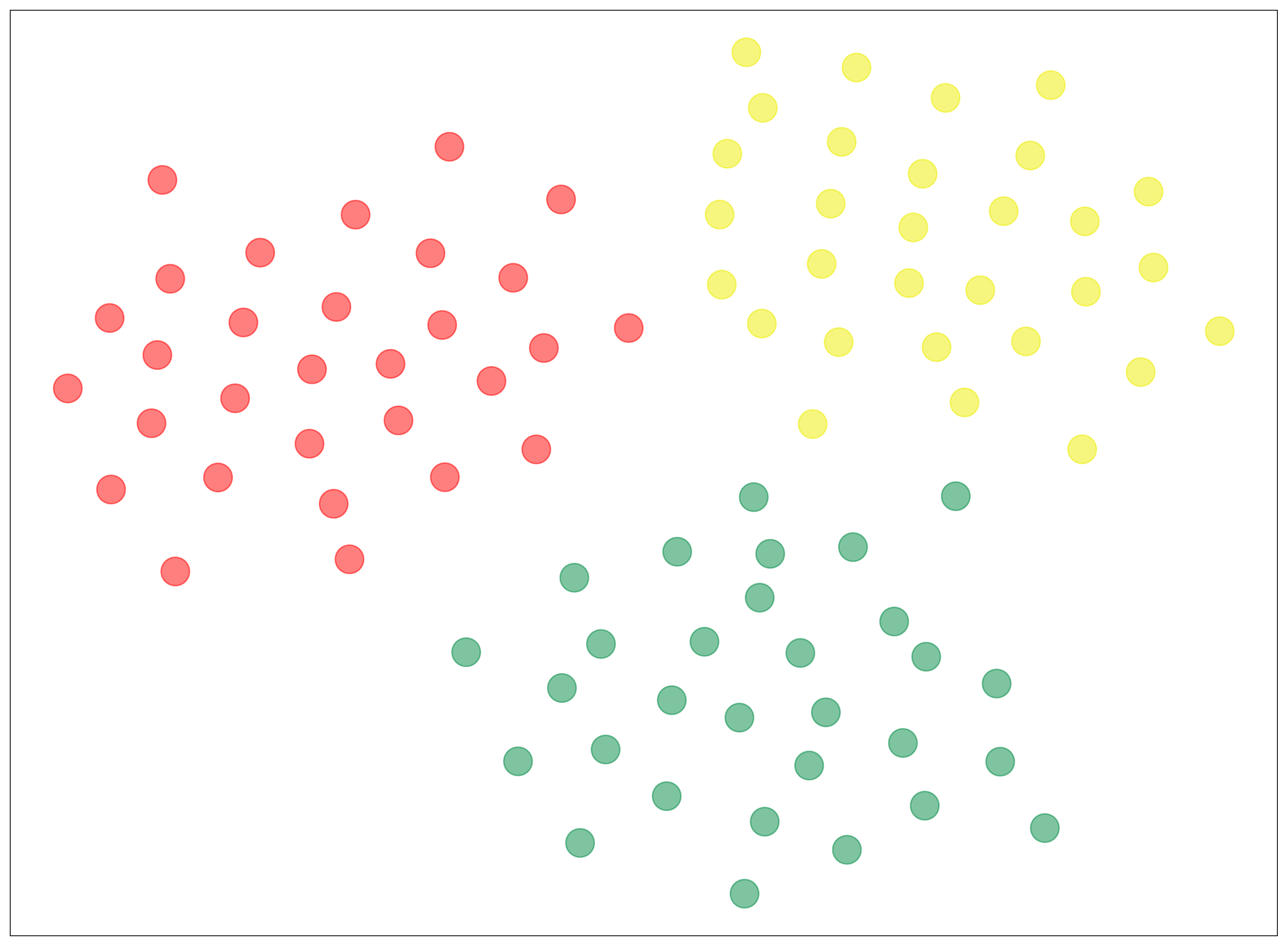}} 
\subfigure[$A_C$]{\includegraphics[width=3.5cm]{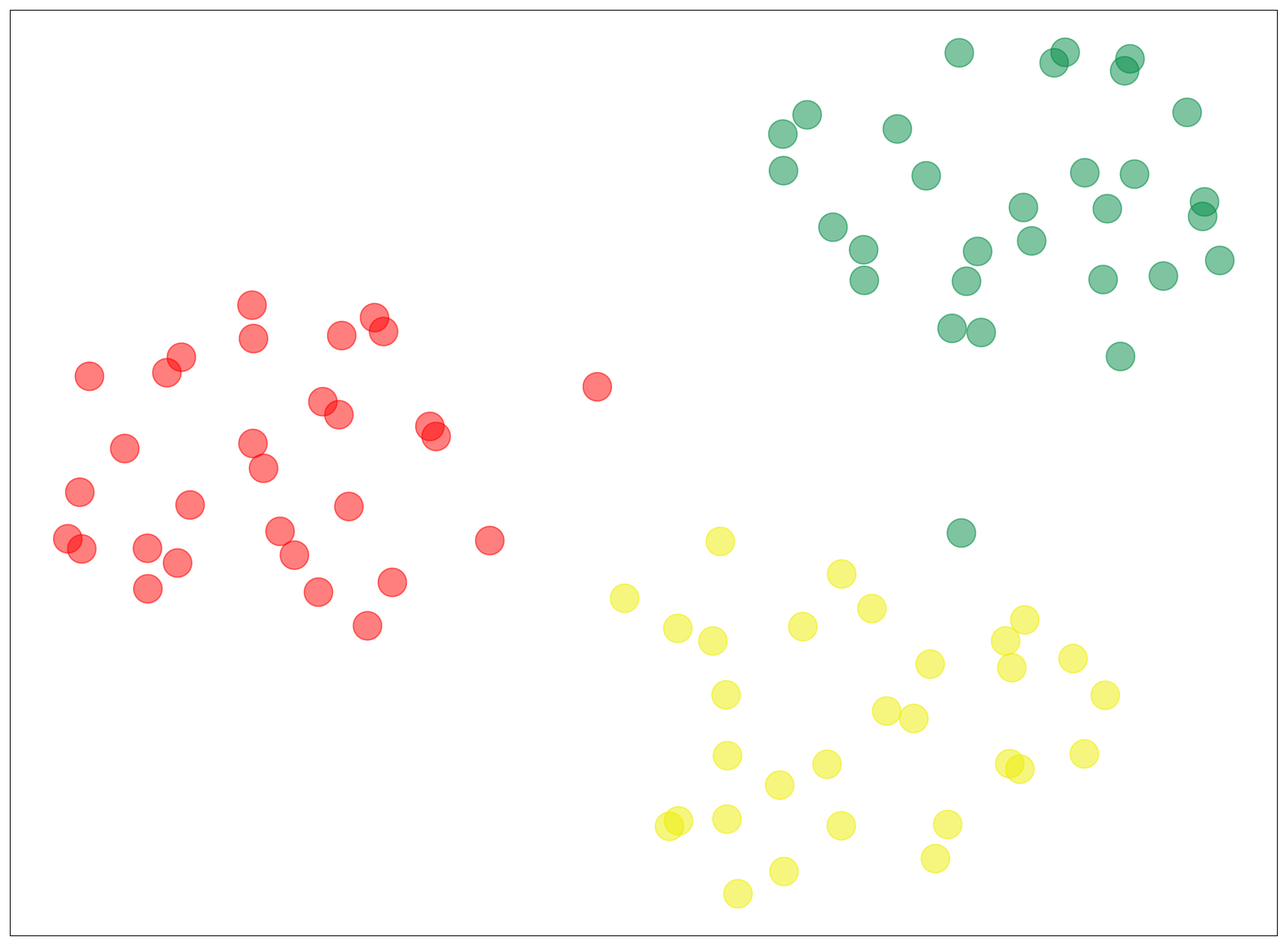}}
\subfigure[$X_C$]{\includegraphics[width=3.5cm]{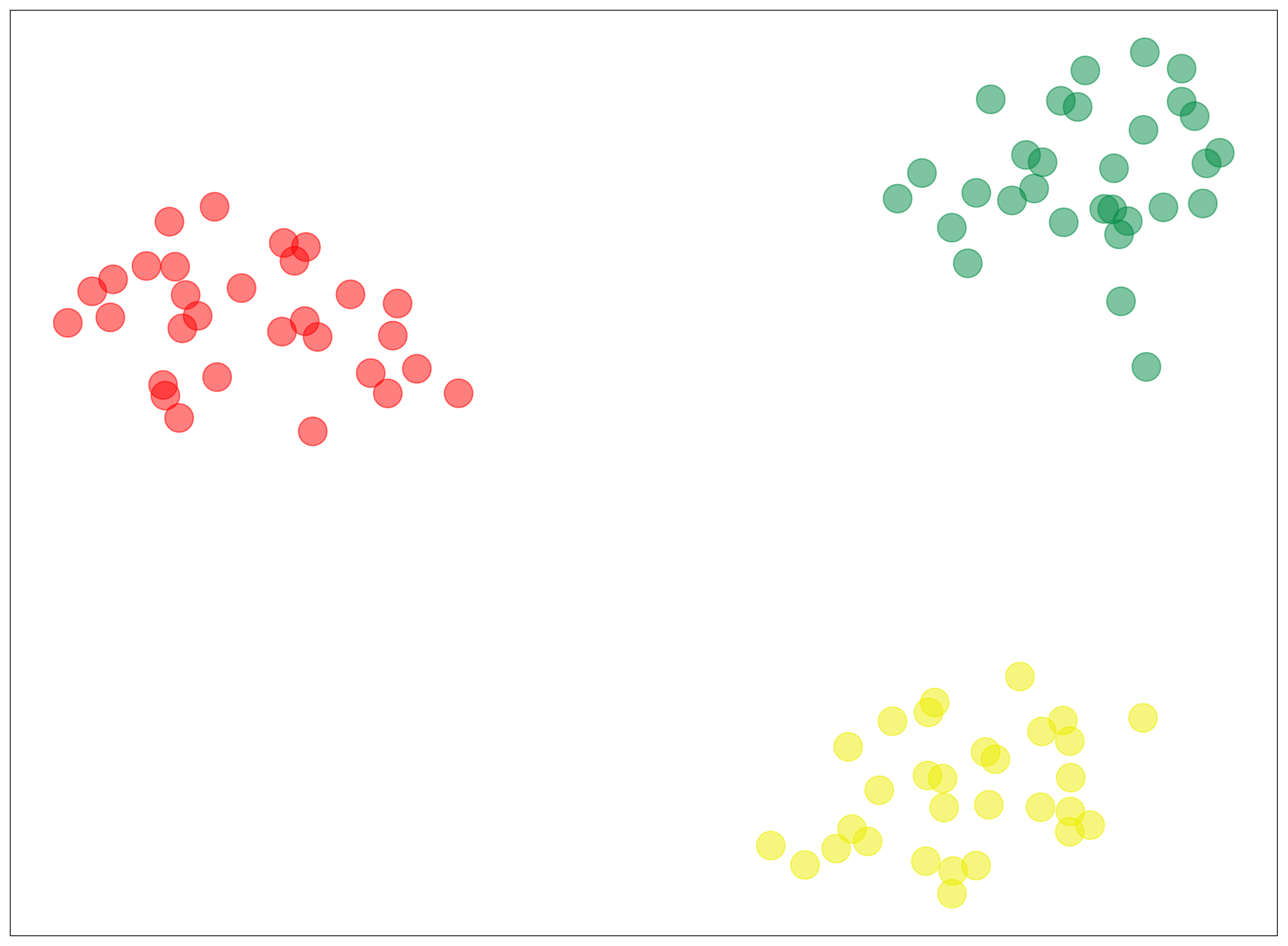}}
\hfill
\centering
\subfigure[A]{\includegraphics[width=3.5cm]{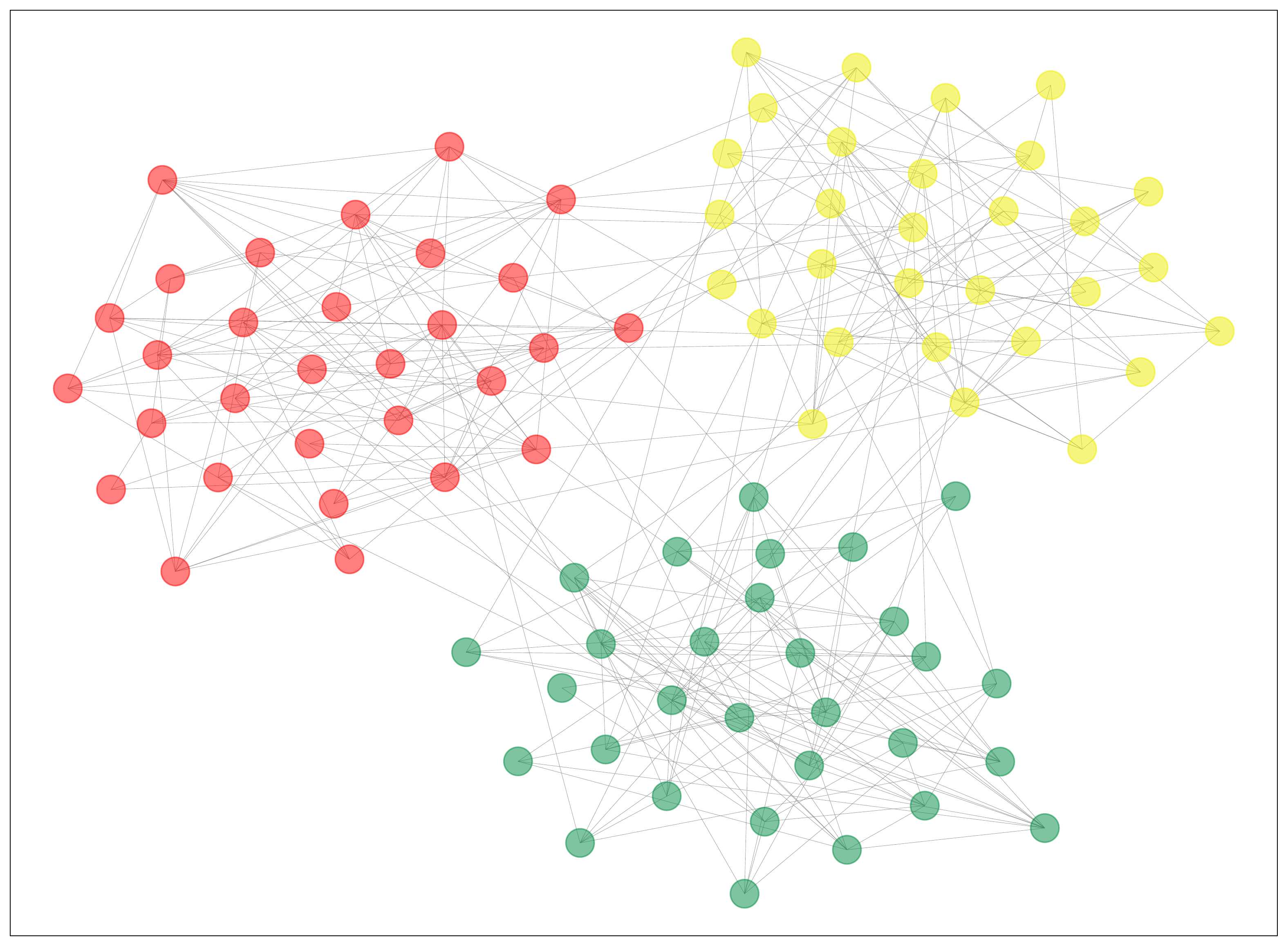}}
\subfigure[$A_T$]{\includegraphics[width=3.5cm]{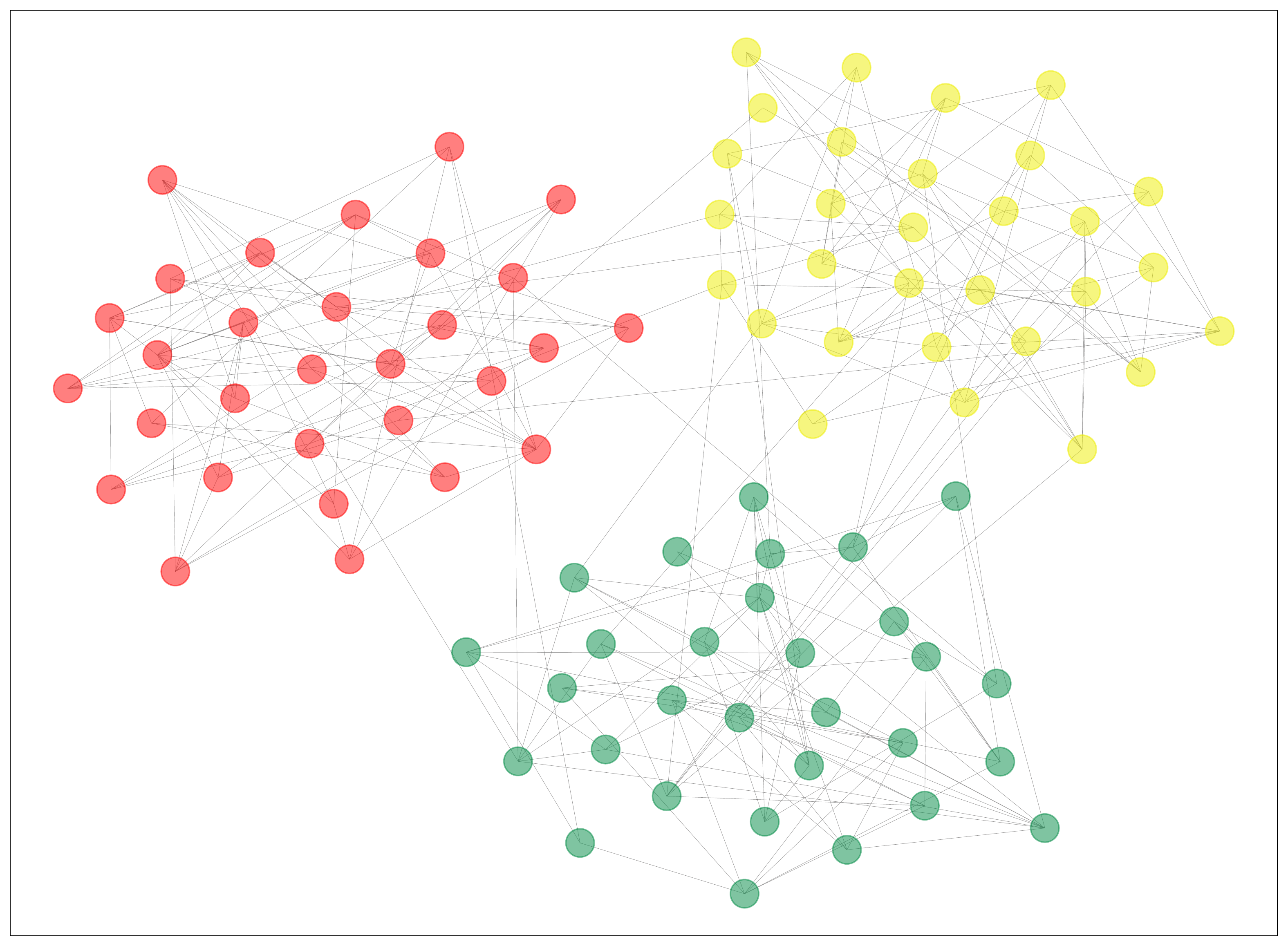}}
\subfigure[$X_T$]{\includegraphics[width=3.5cm]{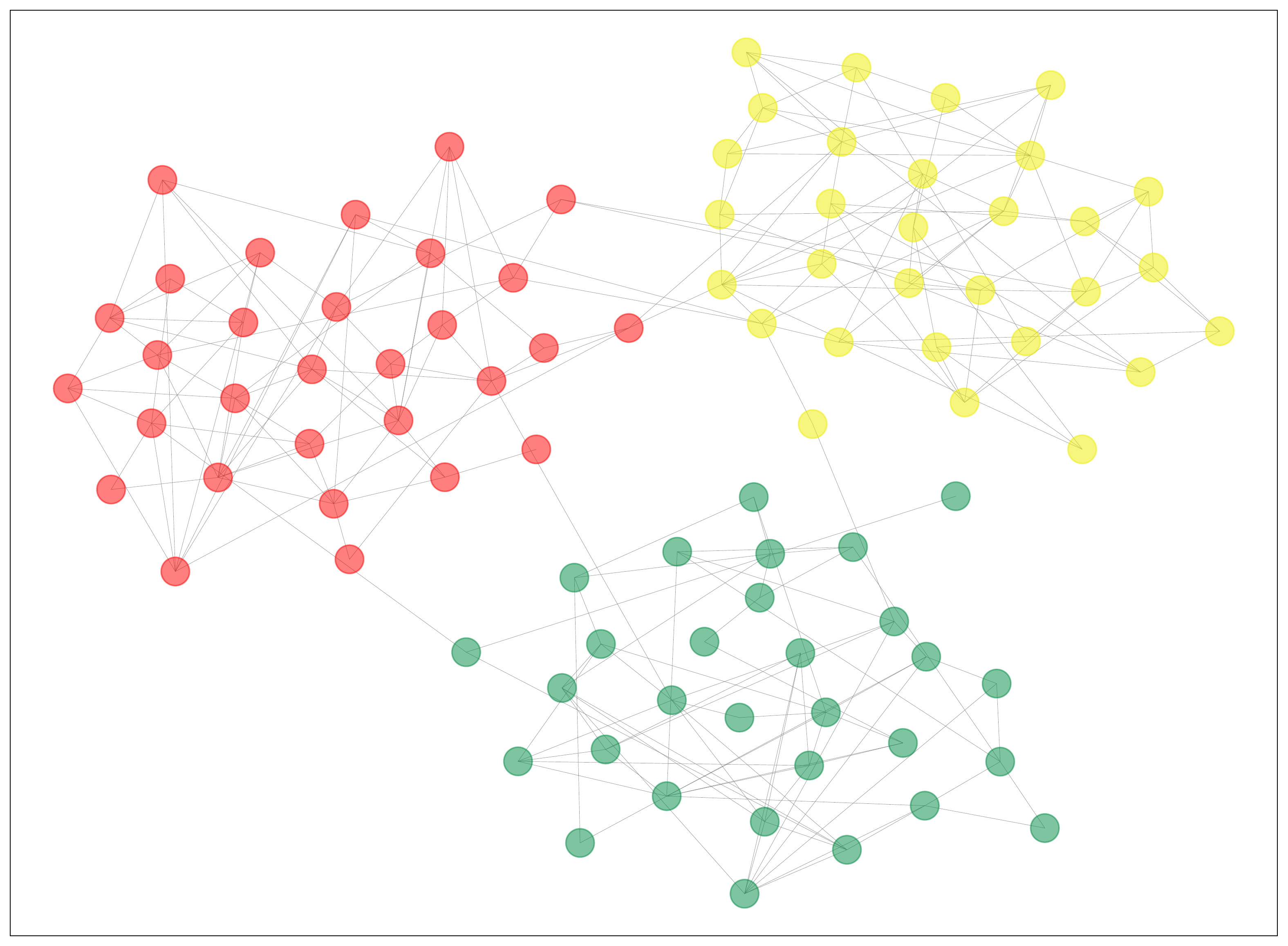}}
\caption{Visualization of attribute features: $X$, $A_C$, and $X_C$, and adjacency features : $A$, $A_T$, and $X_T$.}
\label{adj} 
\end{figure*}

To further investigate the availability of the attribute features $A_C$, $X_C$ and the adjacency features $A_T$, $X_T$, we use a simple yet intuitive case to show the distribution and topology of these augmented features and the original graph feature $A$ and $X$. 
Specifically, we first generate a naive graph consisting of 90 nodes, and randomly assign 3 labels to these nodes.
The edge between every two nodes with the same label is created with the probability of 0.03, and that between every two nodes with different labels is created with the probability of 0.01. Each node has a feature vector of 50 dimensions. We use the Gaussian distribution to generate the node features, the Gaussian distributions for the three classes of nodes have the same covariance matrix, but three different centers far away from each other. 
Then, we can obtain $A$ and $X$ of this graph and augment new features $A_C$, $X_C$, $A_T$, and $X_T$ via the operations described above. As shown in Figure \ref{adj}, the first line shows the node distribution of the attribute features $X$, $A_C$, and $X_C$, we use t-SNE to project them into 2-dimensional spaces. In the second line, we draw edges between nodes via the adjacency features $A$, $A_T$, and $X_T$ to show their different graph topologies, where the node positions are set to be the same as $X$.

\textbf{Attribute Features Analysis.}
The attribute features are $X$, $A_C$, and $X_C$.
First, we can observe that when $X$ is correlated with labels, $X_C$ can preserve the label correlation better, the nodes with the same labels are located in smaller groups and with different labels are farther away from others, we believe that is because $X_C$ preserve the global attribute similarity of nodes with others, and the global information can better improve the node distribution for classification. We can also observe that $A_C$ can preserve the label correlation inherited from $A$, but it presents a totally different node distribution with $X$ as they contain different information. So when the graph topology is correlated with labels and the original attribute $X$ is not, $A_C$ may further improve the accuracy of classification if it is chosen as node attributes.


\textbf{Adjacency Features Analysis.}
The adjacency features are $A$, $A_T$, and $X_T$.
we can observe that comparing with $A$, the topology structure in the augmented feature $A_T$ can preserve the label correlation better, the intra-class connections are denser than the inter-class connections, that may also because  $A_T$ preserve the global structural similarity of nodes with others, and the global information can better improve the graph topology for classification. Also, $X_T$ provide another edge generation method that nodes with the higher similar attribute are more likely to connect each other. So when the node attributes are related with labels and graph topology is not, $X_T$ may further improve the accuracy of classification if it is chosen as the adjacency matrix.

To summarize, the augmented graph features $A_C$, $X_C$, $A_T$, and $X_T$ broaden the availability of the original graph features $X$ and $A$, which is important because the augmentation may improve the distribution of original features for classification by introducing the global information on the one hand, on the other hand, when the distribution of some features are not correlated with the node labels, these information can provide more input choices for GNN model than the original input pair $(A,X)$, and some of them may contribute more than $(A,X)$ for the final task.

\subsection{Attentional Integration Model}
After generating the new inputs for the GNN model, the next question is how do we select useful features.
In the real-world, the graph data is complex, it is hard to know which of the augmented features and original features is correlated with the final task, and time-consuming to manually choose the related ones. So we proposed an attentional integration model, which can automatically assign high weights on features with high correlation for the final task.

Specifically, given the nine combinations of GNN inputs augmented above, we use the traditional GNN encoder, Graph Convolutional Network described in Section 2, to encode the $i$-th inputs into the node embedding matrices $Z_i$:

\begin{equation}
    Z_i = G_i(Adj_i,Att_i) 
\end{equation}

where $Z_i\in \mathbb{R}^{N\times h}$, $h$ is the dimension of output node embedding,  $(Adj_i,Att_i)$ is the $i$-th pairwise input specified in Table \ref{combination}, $G_i(\cdot, \cdot)$ represent the GNN encoder for the $i$-th combination of input, Noted that these nine GNN encoders do not share parameters, this help to better extract the information of different features, but without increasing the time complexity and space complexity because the parameters just increase linearly.
Now we obtain the nine output of node embedding matrices: $\{Z_1,Z_2,...,Z_9\}$ from the nine GNN encoders. Considering they may have different correlations with the node labels, we use an attention mechanism on them to learn their corresponding importance weight and weighted sum them into the final node embedding matrix:

\begin{equation}
    Z = {\alpha}_1\cdot Z_1 + {\alpha}_2\cdot Z_2 + ... +{\alpha}_9\cdot Z_9 
\end{equation}

where $\{\alpha_1, \alpha_2,...,\alpha_9\} \in \mathbb{R}^{N\times 1}$ indicate the attention weights of $n$ nodes with embeddings $\{Z_1,Z_2,...,Z_9\}$, respectively.
To calculate $\alpha_i$, We firstly transform the embeddings through a nonlinear transformation, and then use one shared attention parameter vector $\mathbf{q} \in \mathbb{R}^{h'\times 1}$ to get the attention value $\omega_i$ as follows:

\begin{equation}
    \omega_i = q^T \cdot tanh(W_i \cdot (Z_i)^T + b_i).
\end{equation}

where $\omega_i \in \mathbb{R}^{N\times 1}$, $W_i\in \mathbb{R}^{h'\times h}$ is the weight matrix and $b_i\in \mathbb{R}^{h'\times 1}$ is the bias vector for embedding matrix $Z_i$. 
Then we can get the the attention values $\{\omega_1,\omega_2,...,\omega_9\}$ for embedding matrices $\{Z_1,Z_2,...,Z_9\}$, respectively. We then normalize the attention values $\{\omega_1,\omega_2,...,\omega_9\}$ for each node by softmax function to get the final importance weight:

\begin{equation}
    \alpha_i^j = softmax(\omega_i^j) = \frac{exp(\omega_i^j)}{\sum_{i=1}^9 exp(\omega_i^j)}
\end{equation}

where $\alpha_i^j$ and $\omega_i^j$ represent the $j$-th element of $\alpha_i$ and $\omega_i$, respectively. The larger $\alpha_i^j$ implies the the corresponding node embedding in $Z_i$ is more important for the $j$-th node and should contribute more to its final embedding. 

\subsection{Objective Function}

\subsubsection{Disparity Constraint.}
Firstly, we use the Hilbert-Schmidt Independence Criterion(HSIC)\cite{song2007supervised}, a widely used dependency measurement\cite{zhang2018fish,ma2020hsic}, as a penalty term in the objective function to ensure the nine output node embeddings $\{Z_1,Z_2,...,Z_9\}$ encoded from nine inputs can capture non-redundant information.
HSIC is simple and reliable to compute the independency between variables and the smaller the value is, the more independent they are. The HISC of any two embeddings $Z_i$ and $Z_j$ is defined as:

\begin{equation}
    HSIC(Z_i,Z_j) = (n-1)^{-2}tr(K_i H K_j H),
\end{equation}

where $K_i, K_j\in R^{N\times N}$ are the Gram matrices with ${K_i}^{uv} = k_i({Z_i}^u,{Z_i}^v),{K_j}^{uv} = k_j({Z_j}^u,{Z_j}^v)$, ${K_i}^{uv}$ is the element in $u$-th row and $v$-th column of $K_i$, ${Z_i}^u$ is the $u$-th row of $Z_i$, and $k_i(\cdot, \cdot)$ is the kernel function.
$H = I - n^{-1}ee^T$, where $e$ is an all-one column vector and $I$ is an identity matrix. In our implementation, we use the inner product kernel function. 
Then we set the disparity constraint $\mathcal{L}_d$ by minimizing the values of HISC among nine output nodes embeddings:
\begin{equation}
    \mathcal{L}_d = \sum_{i \neq j} HISC(Z_i,Z_j).
\end{equation}


\subsubsection{Optimization Objective.}
For semi-supervised multi-class classification, We feed the final node embeddings $Z$ into a linear transformation and a $softmax$ function. Denote classes set is $C$, and the probability of node $i$ belonging to class $c \in C$ is $\hat{Y}_{ic}$, the prediction results on whole nodes  $\hat{Y} = [\hat{Y}_{ic}]\in \mathbb{R}^{N\times C}$ can be calculated as:

\begin{equation}
    \hat{Y} = softmax(W\cdot Z + b),
\end{equation}

where $softmax(x)=\frac{exp(x)}{\sum_{c=1}^C exp(x_c)}$ is actually a row-wise normalizer across all classes. Then the cross-entropy loss $\mathcal{L}$ for node classification over all labeled nodes is represented as:

\begin{equation}
    \mathcal{L}_l = -\sum_{l\in \mathcal{Y}_L}\sum_{c=l}^C Y_{lc}ln\hat{Y}_{lc}.
\end{equation}

Where $\mathcal{Y}_L$ is the set of node indices that have labels, for each $l\in L$ the real one-hot encoded label is $Y_l$. 

Finally, combining the node classification task and the disparity constraints, we have the following overall objective function:

\begin{equation}
    \mathcal{L} = \mathcal{L}_l + \lambda\mathcal{L}_d.
\end{equation}

where $\lambda$ is parameters of the disparity constraint terms. We use a mini-batch Adam optimizer to minimize $\mathcal{L}$ and optimize the parameters in the whole model. 
Noted that we use HISC to calculate the pairwise independence, it would take $C_9^2$ times of calculation of HISC among $Z_1$ to $Z_9$ in each training step, which we think is unnecessary. We use a sampling strategy to reduce the computation that randomly selecting $t$ pairs of the output embeddings and summing their HISC as the disparity constraints loss in each training step. Through multiple iterations, all combinations of embeddings should be sampled and all embeddings should be trained to be independent of each other.

\begin{table*}[]
    \centering
    \caption{The statistics of the datasets}
        \begin{tabular}{lcccc}
        \toprule
        Dataset     & Nodes & Edges  & Classes & Attribute \\ \midrule
        Citeseer    & 3327  & 4732   & 6       & 3703     \\
        UAI2010     & 3067  & 28311  & 19      & 4973     \\
        ACM         & 3025  & 13128  & 3       & 1870     \\
        BlogCatalog & 5196  & 171743 & 6       & 8189     \\
        Flickr      & 7575  & 239738 & 9       & 12047
        \\ \bottomrule

        \end{tabular}
    \label{statistics}
\end{table*}

\section{Experiments}

\subsection{Experiment Setting}
To adequately examine the effectiveness of our proposed data augmentation method, we evaluate the performance of our framework on five real-world benchmark datasets:
Citeseer\cite{kipf2016semi} is research paper citation network, 
UAI2010\cite{wang2018unified} is a dataset for community detection, 
ACM\cite{wang2019heterogeneous} is research paper coauthor network extracted from ACM dataset, 
BlogCatalog\cite{meng2019co} is a social network with bloggers relationships extracted from the BlogCatalog website,
Flickr\cite{meng2019co}is a social network with users interaction from an image and video hosting website.
Basic statistics of these datasets are summarized in Table \ref{statistics}.

We compared our method with some GCN and node classification related baselines: GCN\cite{kipf2016semi} is a classical semi-supervised graph convolutional network model, which obtains node representation through multi-layer neighbor aggregation.
Chebyshev\cite{defferrard2016convolutional} learns rich feature information by superimposing multiple Chebyshev filters with GCN. 
GAT\cite{velivckovic2017graph} is a graph neural network model that aggregates node features through multiple attention heads with different semantics. 
DEMO-Net\cite{wu2019net} proposes a generic graph neural network model which formulates the feature aggregation into a multi-task learning problem according to nodes' degree values. 
MixHop\cite{abu2019mixhop} utilizes multiple powers of the adjacency matrix to learn the general mixing of neighborhood information, including averaging and delta operators in the feature space.
We also compare our method with some related graph data augmentation based methods for semi-supervised node classification.
GAug\cite{zhao2020data} is to leverage information inherent in the graph to predict which non-existent edges should likely exist, and which existent edges should likely be removed in the original graph to produce modified graphs to improve the model performance.
MCGL\cite{dong2020data} assigns pseudo-labels to some nodes in each convolutional layer, and improves the performance of the model by expanding the training set.

The weights of parameters are initialized like the original GCN\cite{kipf2016semi} and input vectors are row-normalized accordingly\cite{glorot2010understanding}. For our model, we train nine 2-layer GCNs with the same hidden layer dimension($h_1$) and the same output dimension ($h_2$) simultaneously, where $h_1$ of the UAI2010, BlogCatalog, and Flickr is 256 and the out dimension $h_2$ is 128. The $h_1$ and $h_2$ of ACM and Citeseer are 512 and 256 respectively. we use $5e-4$ learning rate with Adam optimizer, the dropout rate is 0.5, weight decay is $1e-4$. In addition, the hyper-parameter $k$ for constructing $k$-nearest neighbor graphs is 4, $t$ for sampling embeddings pairs is 8. 
For the baselines, we set the dimension of node embeddings in five datasets same as the setting of out method, and  the other hyper-parameter setting are based on default values or the values specified in their own papers.
We choose the number of labeled nodes per class as 20/40/60 respectively for training, and 500 nodes are used for validation and 1000 nodes for testing. 
All methods are repeatedly run 5 times, the average results are reported to make sure the results can reflect the performances of methods.

\subsection{Semi-Supervised Classification}
The semi-supervised node classification results are reported in Table \ref{classification}. We report the Accuracy (ACC) and macro F1-score (F1) of the classification results. From the results, we can observe that (1) our proposed method achieves the best performance on all datasets with all label rates, showing the superiority of our method in improving the semi-supervised node classification. (2) Our method consistently outperform the original GCN on all five datasets, the improvement of ACC over Citeseer, UAI2010, ACM, BlogCatalog, Flickr is \{3.0\%-6.1\%,41.3\%-44.9\%,2.5\%-4.6\%,20.4\%-25.1\%,71.5\%-84.2\%\}, respectively. indicating that the augmented graph features contain more useful information than original graph features and help to node classification.
(3) We noticed that two graph augmentation methods GAug and MCGL perform well on some datasets, but also fail in some datasets, while our method consistently performs well on all datasets, showing that our whole framework is robust on different types of graphs.

\begin{figure}[htbp]
\setlength{\belowcaptionskip}{-0.5cm} 
\centering
\subfigure[GCN(Citeseer)]{
\label{fig:a} 
\begin{minipage}[t]{0.32\linewidth}
\centering
\includegraphics[width=1\linewidth]{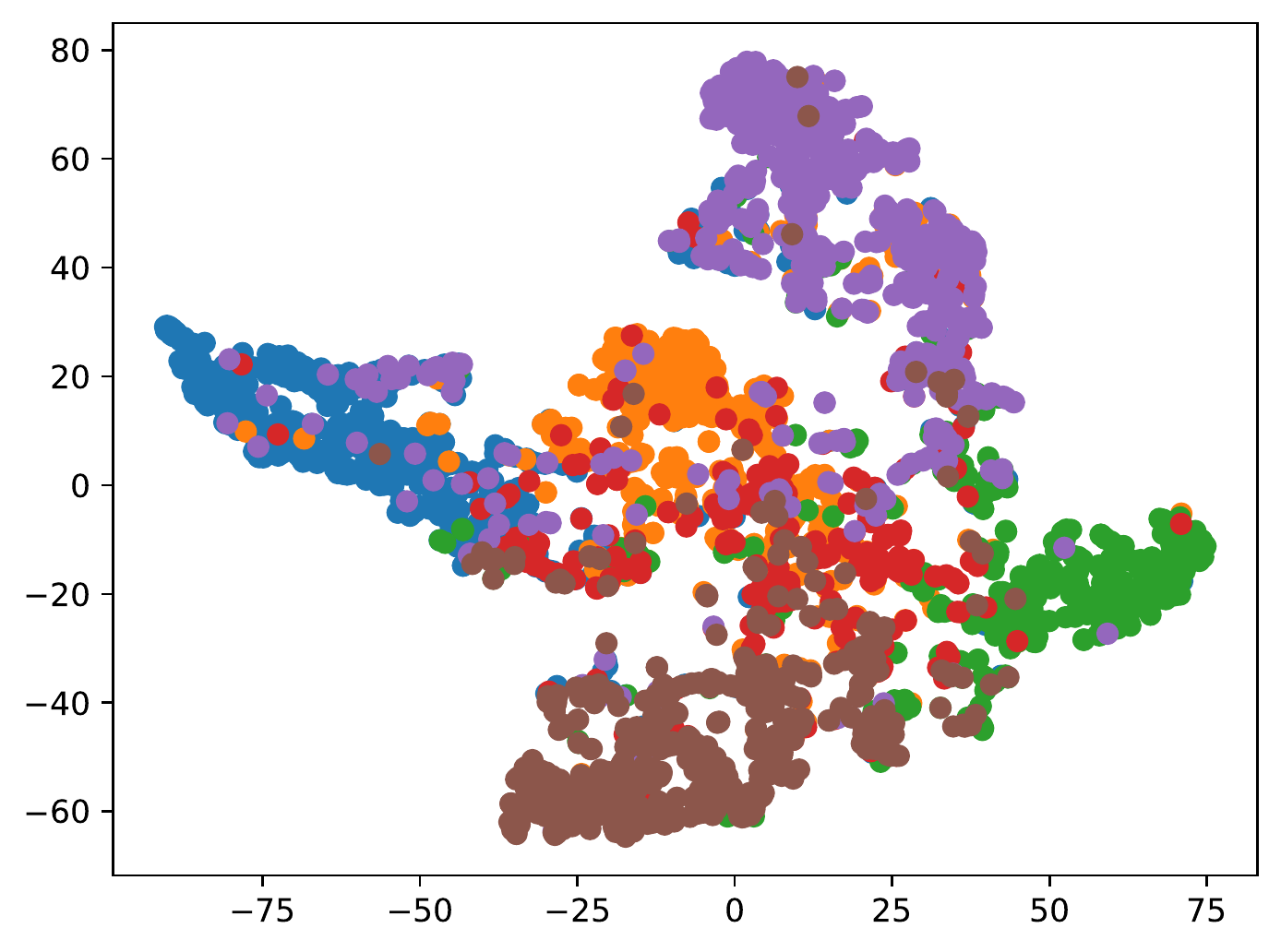}
\end{minipage}%
}%
\subfigure[GCN(UAI2010)]{
\label{fig:a} 
\begin{minipage}[t]{0.32\linewidth}
\centering
\includegraphics[width=1\linewidth]{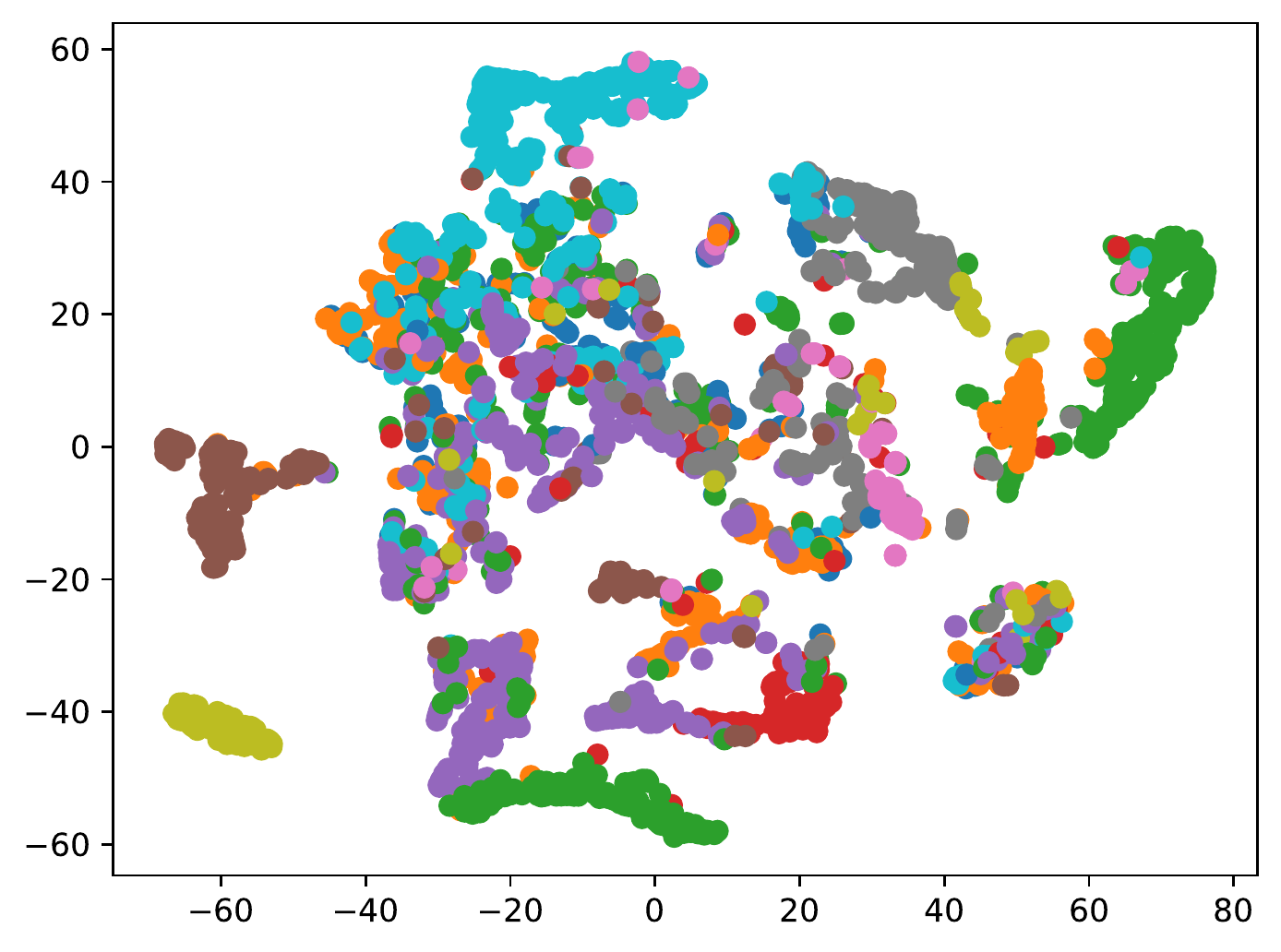}
\end{minipage}%
}%
\subfigure[GCN(ACM)]{
\label{fig:b} 
\begin{minipage}[t]{0.32\linewidth}
\centering
\includegraphics[width=1\linewidth]{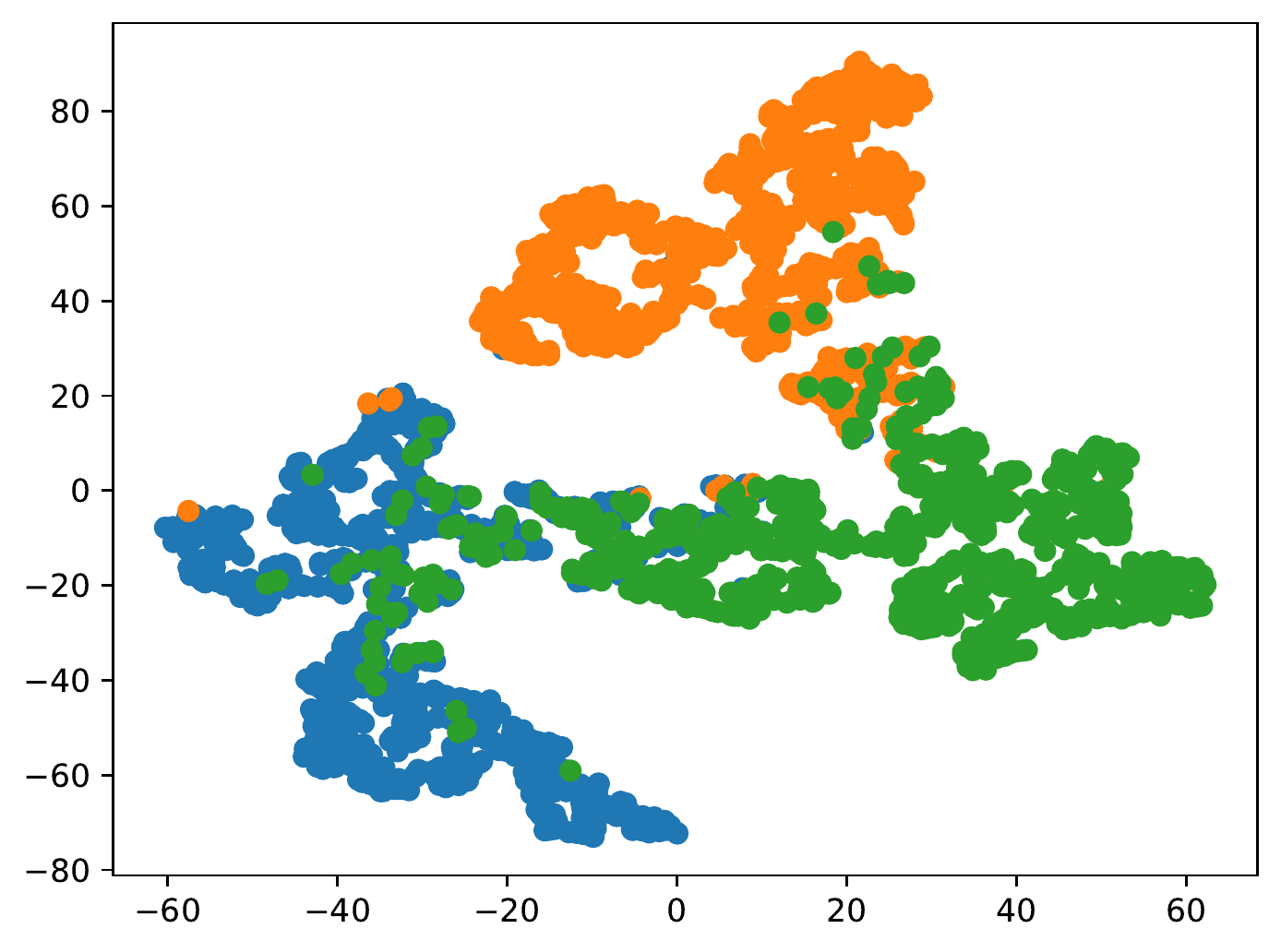}
\end{minipage}%
}%
\hfill

\subfigure[Ours(Citeseer)]{
\label{fig:a} 
\begin{minipage}[t]{0.32\linewidth}
\centering
\includegraphics[width=1\linewidth]{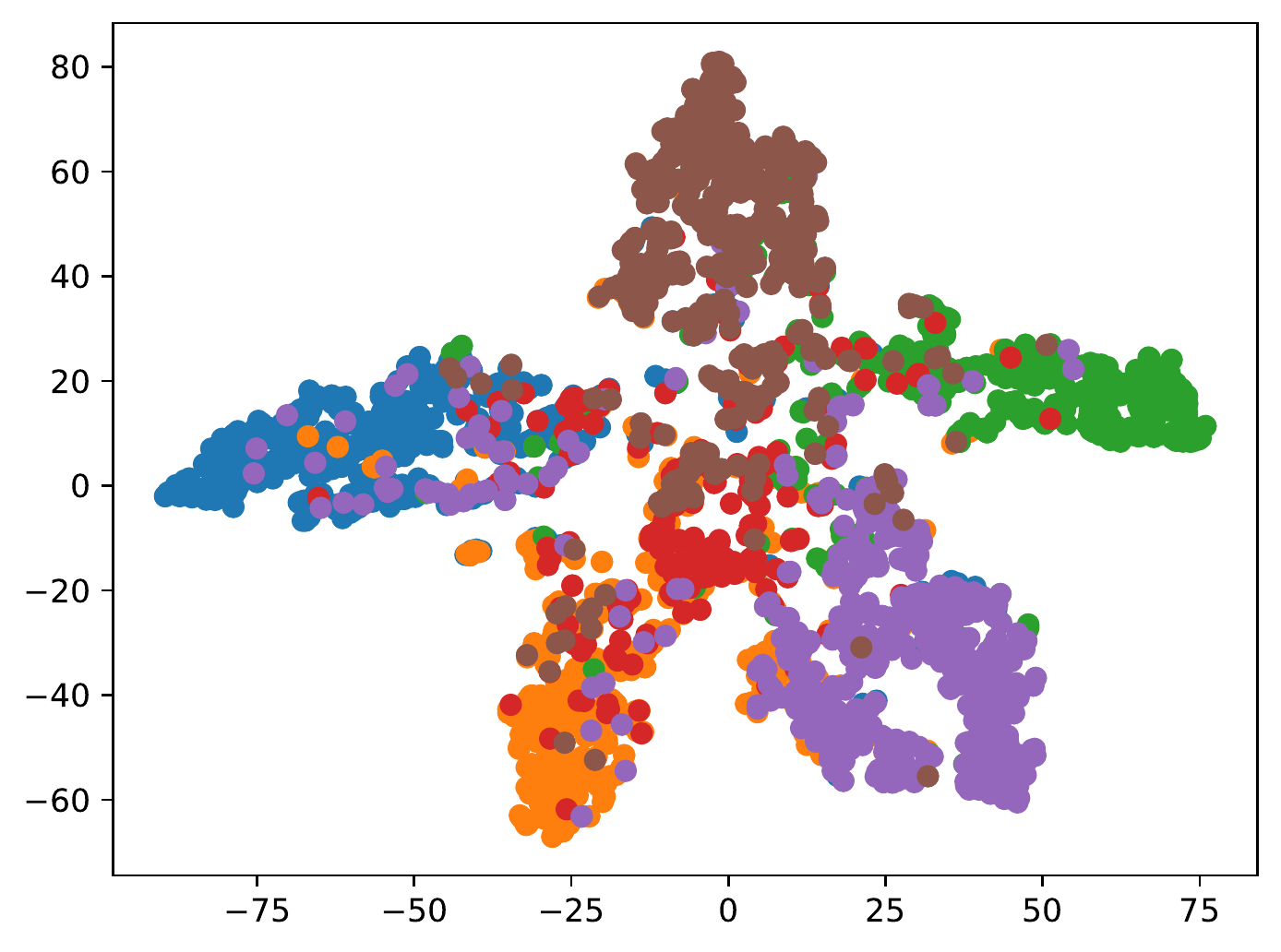}
\end{minipage}%
}%
\subfigure[Ours(UAI2010)]{
\label{fig:a} 
\begin{minipage}[t]{0.32\linewidth}
\centering
\includegraphics[width=1\linewidth]{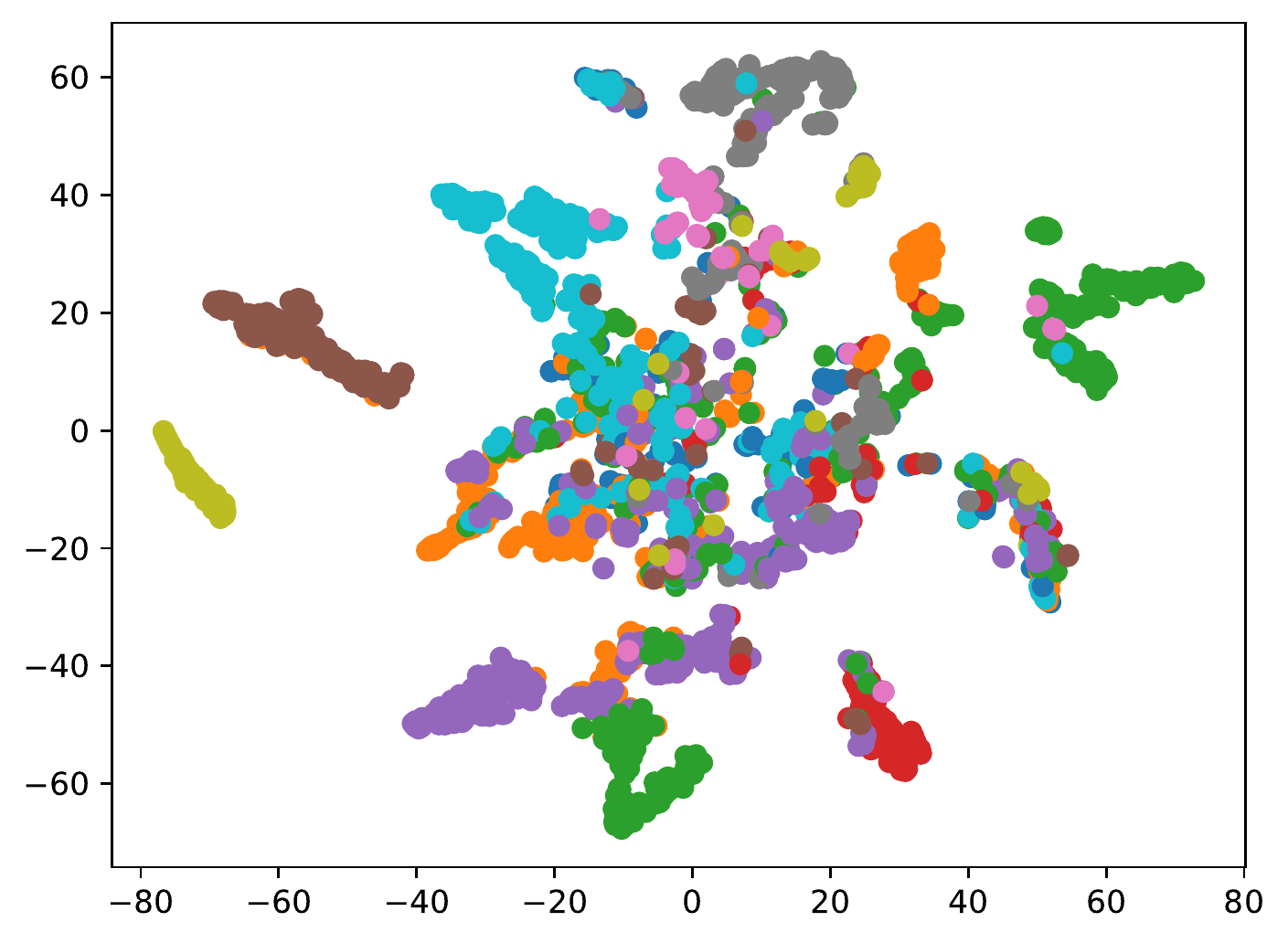}
\end{minipage}%
}%
\subfigure[Ours(ACM)]{
\label{fig:b} 
\begin{minipage}[t]{0.32\linewidth}
\centering
\includegraphics[width=1\linewidth]{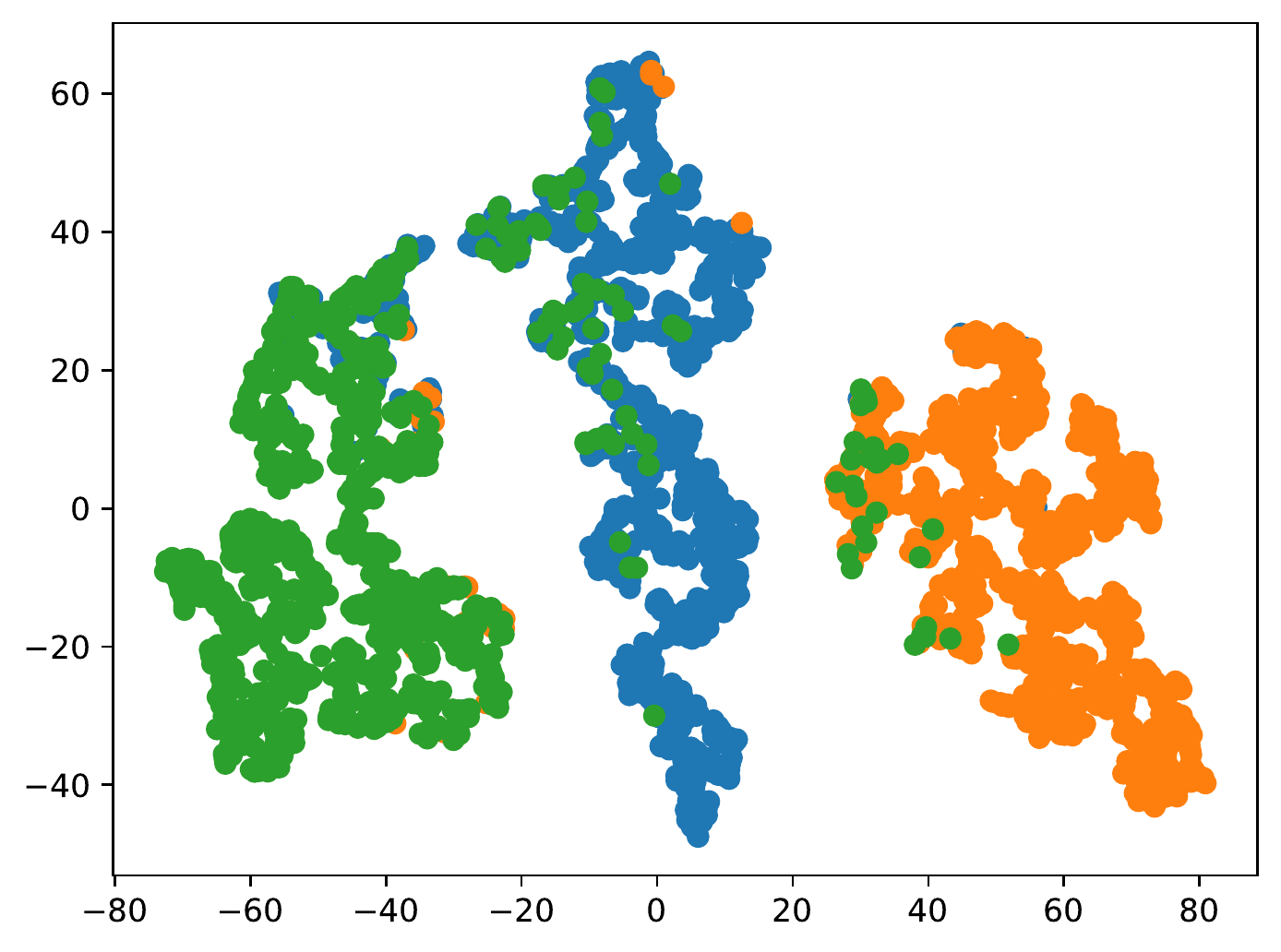}
\end{minipage}%
}%
\centering
\caption{Visualization of the learned final node embeddings on ACM, UAI2010, and Citeseer datasets. (L/C=20)}
\label{cluster} 
\end{figure}

We further report the visualization of learned node embeddings of the Citeseer, UAI2010, and ACM datasets in Figure \ref{cluster}. We use t-SNE to project the final node embeddings of our method and original GCN into 2-dimensional spaces and color nodes differently according to their labels. We can observe that the boundaries between different classes in our method are sharper than the original GCN, and nodes in the same class are more concentrated, especially in the Citeseer dataset, which proves our method can learn better node representations to improves the node classification performance of original GCN.

\begin{table*}[htbp]
\centering
\caption{Results of semi-supervised node classification(\%). (Bold: best. L/C is the number of labeled nodes per class. The results of some baselines are taken from \cite{Wang2020AM}.) }
\begin{tabular}{c|c|cc|cc|cc|cc|cc}
\toprule
\multicolumn{2}{c|}{Datasets}     & \multicolumn{2}{c|}{Citeseer}   & \multicolumn{2}{c|}{UAI2010}     & \multicolumn{2}{c|}{ACM}         & \multicolumn{2}{c|}{BlogCatalog} & \multicolumn{2}{c}{Flickr} \\ \midrule
L/C & Method & ACC   & \multicolumn{1}{c|}{F1} & ACC   & \multicolumn{1}{c|}{F1} & ACC   & \multicolumn{1}{c|}{F1} & ACC   & \multicolumn{1}{c|}{F1} & ACC          & F1          \\ \midrule
\multirow{8}{*}{20}      
    &  GCN  & 70.30  & 67.50  & 49.88  & 32.86 & 87.80  & 87.82  & 69.84  & 68.73  & 41.42   & 39.95   \\
    & Chebyshev    &  69.80 & 65.92 & 50.02 & 33.65  & 75.24 & 74.86 & 38.08 & 33.39 & 23.26 & 21.27 \\
    &  GAT  & 72.50  & 68.14  & 56.92  & 39.61 & 87.36  & 87.44  & 64.08  & 63.38  & 38.52   & 37.00   \\
    &  DEMO-Net  & 69.50  & 67.84  & 23.45  & 16.82 & 84.48  & 84.16  & 54.19  & 52.79  & 34.89   & 33.53  \\
    &  MixHop  & 71.40  & 66.96  & 61.56  & 49.19 & 81.08  & 81.40  & 65.46  & 64.89  & 39.56   & 40.13   \\
    &  GAug  & 73.30  & 70.12   & 52.96   & 49.82  & 90.82   & 89.44   & 77.60   & 75.43   & 68.20    & 67.55    \\
    &  MCGL  & 66.88   & 63.26   & 42.56   & 24.78  & 90.95   & 91.01   & 54.22   & 50.15   & 15.67    & 15.54    \\
    &  Ours  & \textbf{74.60}  & \textbf{70.20}  & \textbf{72.20}  & \textbf{60.87} & \textbf{91.90}  & \textbf{91.81}  & \textbf{84.10}  & \textbf{84.60}  & \textbf{76.30}   & \textbf{76.27}   \\ \midrule
\multirow{8}{*}{40}      
    &  GCN  & 73.10  & 69.70  & 51.80  & 33.80 & 89.06  & 89.00  & 71.28  & 70.71  & 45.48   & 43.27   \\
    & Chebyshev    &  71.64 & 68.31 & 58.18 & 38.80  & 81.64 & 81.26 & 56.28 & 53.86 & 35.10 & 33.53 \\
    &  GAT  & 73.04  & 69.58  & 63.74  & 45.08 & 88.60  & 88.55  & 67.40  & 66.39  & 38.44   & 39.94   \\
    &  DEMO-Net  & 70.44  & 66.97  & 30.29  & 26.36 & 85.70  & 84.83  & 63.47  & 63.09  & 46.57   & 45.23   \\
    &  MixHop  & 71.48  & 67.40  & 65.05  & 53.86 & 82.34  & 81.13  & 71.66  & 70.84  & 55.19   & 56.25   \\
    &  GAug  & 74.60   & 71.32   & 55.26   & 53.36  & 91.24   & 91.01   & 79.46   & 77.79   & 73.24    & 72.28    \\
    &  MCGL  & 69.48   & 65.98   & 41.93   & 25.72  & 91.10   & 91.13   & 54.74   & 51.24   & 17.82    & 17.06   \\
    &  Ours  & \textbf{75.50}  & \textbf{71.58}  & \textbf{75.10}  & \textbf{69.70} & \textbf{92.10}  & \textbf{91.94}  & \textbf{89.20}  & \textbf{89.06}  & \textbf{80.10}   & \textbf{79.36}   \\ \midrule
\multirow{8}{*}{60}      
    &  GCN  & 74.48  & 71.24  & 54.40 & 34.12 & 90.54  & 90.49  & 72.66  & 71.80  & 47.96   & 46.58   \\
    & Chebyshev    &  73.26 & 70.31 & 59.82 & 40.60  & 85.43 & 85.26 & 70.06 & 68.37 & 41.70 & 40.17 \\
    &  GAT  & 74.76  & 71.60  & 68.44  & 48.97 & 90.40  & 90.39  & 69.95  & 69.08  & 38.96   & 37.35   \\
    &  DEMO-Net  & 71.86  & 68.22  & 34.11  & 29.05 & 86.55  & 84.05  & 76.81  & 76.73  & 57.30   & 56.49   \\
    &  MixHop  & 72.16  & 69.31  & 67.66  & 56.31 & 83.09  & 82.24  & 77.44  & 76.38  & 64.96   & 65.73   \\
    &  GAug  & 75.48   & 72.22   & 55.92   & 54.08  & 92.06   & 91.81   & 81.81   & 79.84   & 75.68   & 74.24    \\
    &  MCGL  & 74.02   & 70.69   & 44.30   & 22.46  & 92.03   & 92.04   & 55.24   & 49.41   & 22.36    & 21.28    \\
    &  Ours  & \textbf{76.70}  & \textbf{72.88}  & \textbf{76.90}  & \textbf{69.79} & \textbf{92.80}  & \textbf{92.75}  & \textbf{89.70}  & \textbf{89.53}  & \textbf{82.29}   & \textbf{82.85}   \\ \bottomrule
\end{tabular}
\label{classification}
\end{table*}
\vspace{-0.5cm}

\subsection{Attentional Integration Model Analysis}

\begin{figure}[htbp]
\centering
\subfigure[Citeseer]{
\label{fig:a} 
\begin{minipage}[t]{0.31\linewidth}
\centering
\includegraphics[width=1\linewidth]{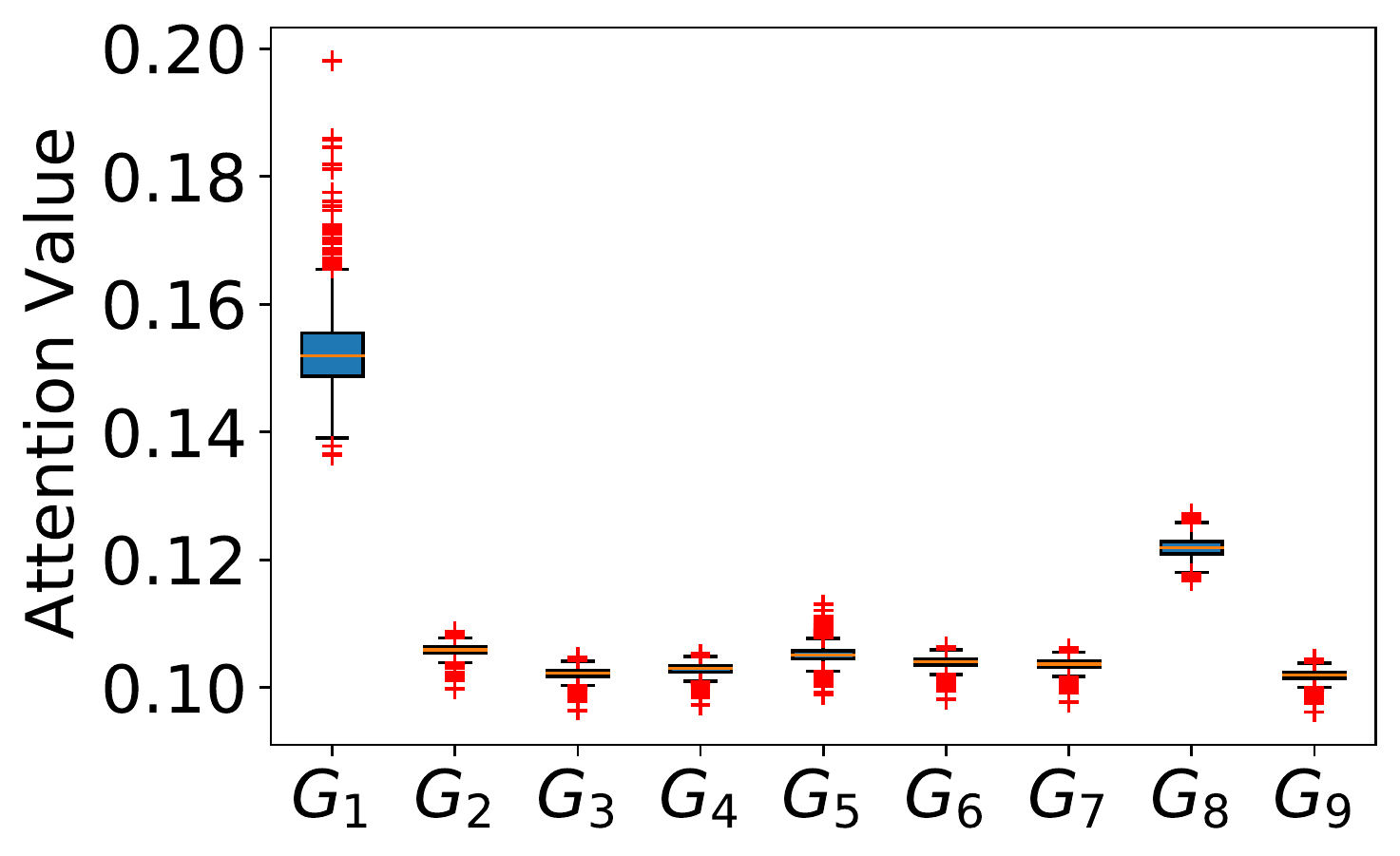}
\end{minipage}%
}%
\subfigure[UAI2010]{
\label{fig:b} 
\begin{minipage}[t]{0.31\linewidth}
\centering
\includegraphics[width=1\linewidth]{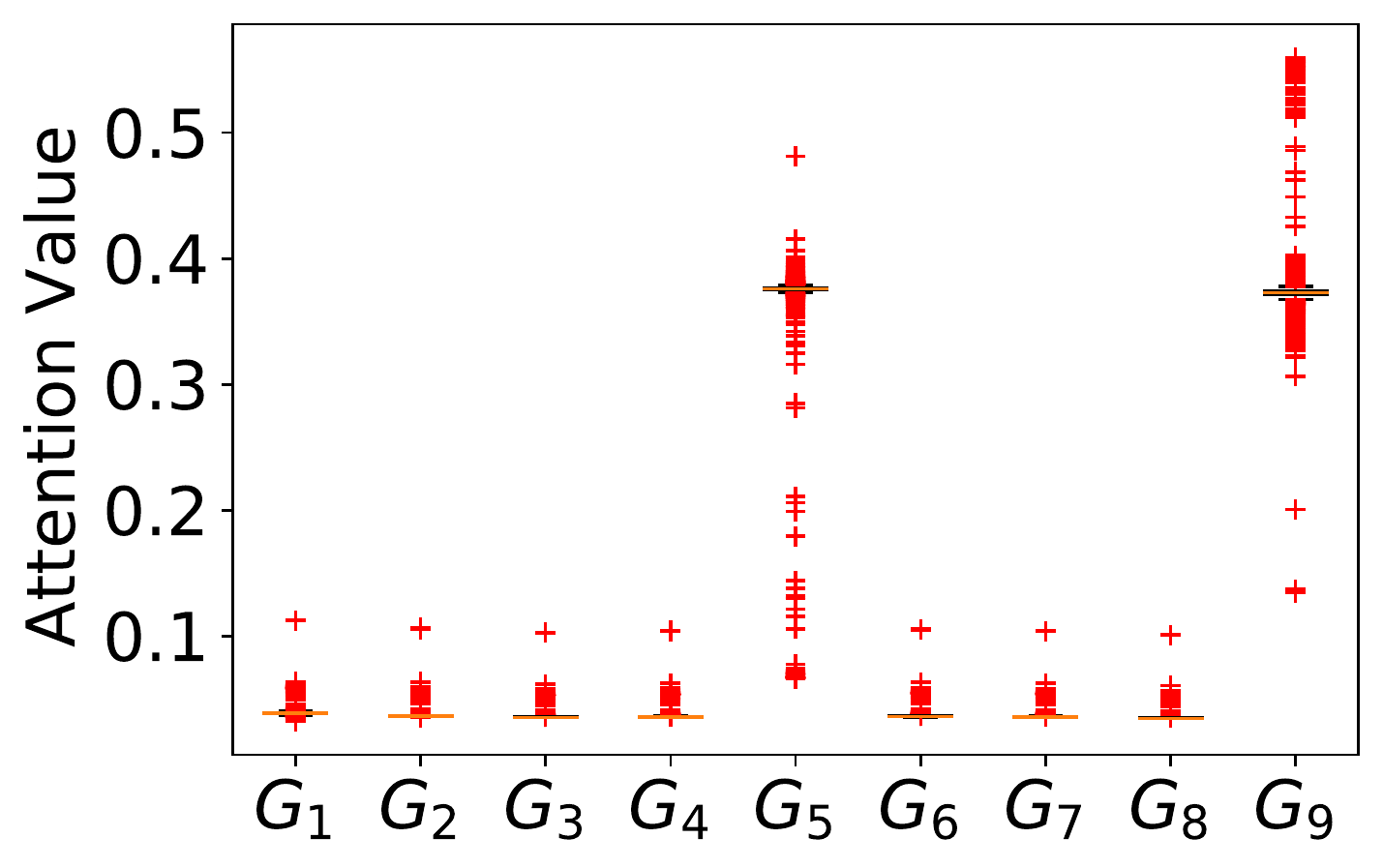}
\end{minipage}%
}%
\subfigure[ACM]{
\label{fig:c} 
\begin{minipage}[t]{0.315\linewidth}
\centering
\includegraphics[width=1\linewidth]{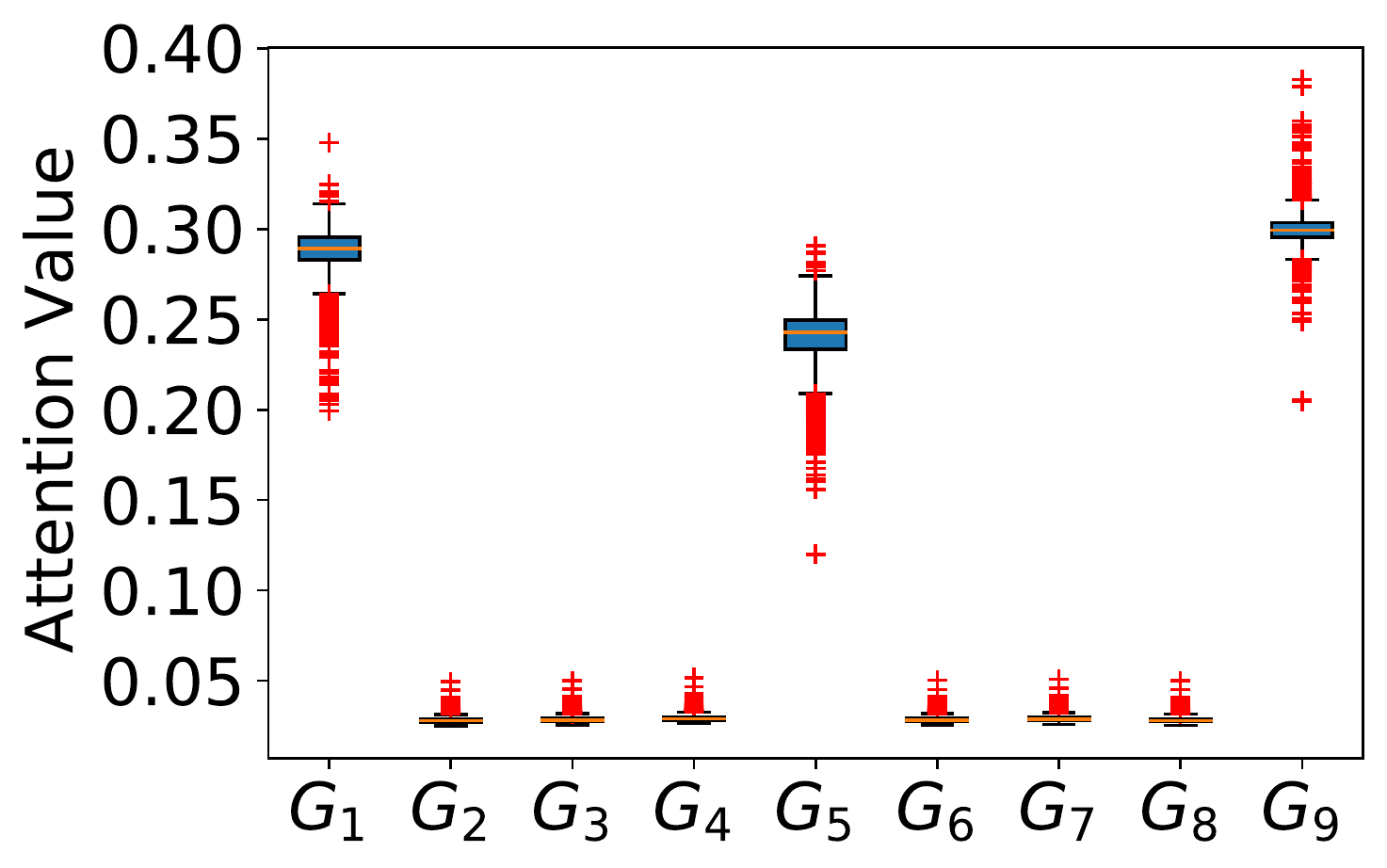}
\end{minipage}%
}%
\centering
\caption{Analysis of attention distribution. (L/C=20)}
\label{attention_max} 
\end{figure}

We design nine combinations of features as inputs of GCN models and learn nine specific node embeddings for each node, 
then each embedding is associated with the corresponding attention values by our proposed attentional integrating model. 
Thus, we conduct attention distribution analysis on the ACM, UAI2010, and Citeseer datasets in Figure \ref{attention_max}, we report the Box-plots of the learned attention value distributions of all nodes respectively for nine GCN models $\{G_1,...,G_9\}$.
We can observe that the average of attention values for nine input combinations are evidently different, some of the combinations may have larger attention values than others,
For example in ACM, the attention values of $G_1$, $G_5$, and $G_9$ are larger than others, which implies that the corresponding augmented inputs of $(A,X)$, $(A_T,A_C)$, and $(X_T,X_C)$ contain more valuable information than other inputs for the classification task. Also, we can observe that between different datasets, the same combination input may be quite different in attention values, which proves that our proposed attentional integrating model is able to adaptively find and assign larger attention value for the important information on different datasets.   

\begin{figure}[htbp]
\centering
\subfigure[Citeseer]{
\label{fig:a} 
\begin{minipage}[t]{0.307\linewidth}
\centering
\includegraphics[width=1\linewidth]{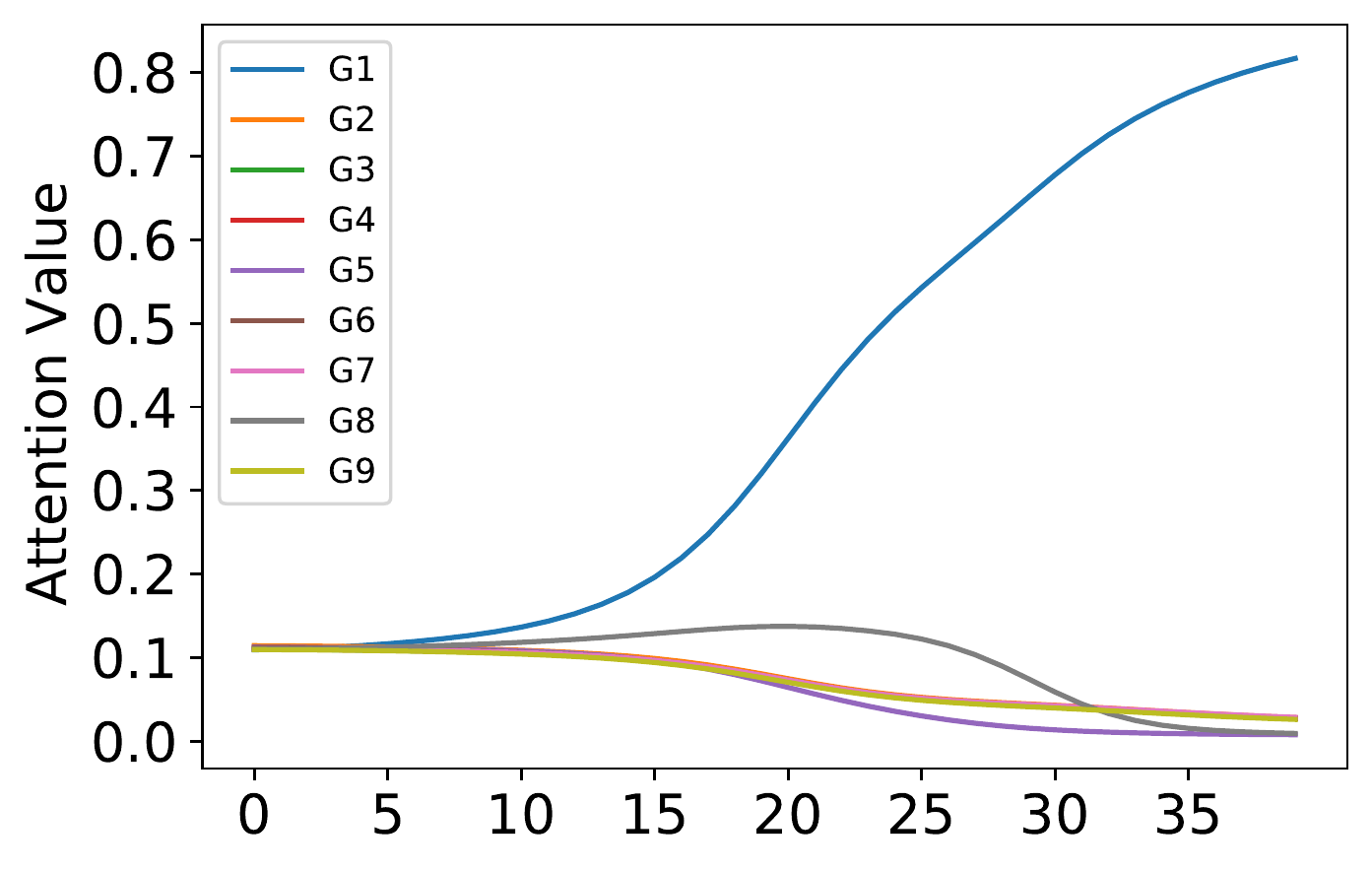}
\end{minipage}%
}%
\subfigure[UAI2010]{
\label{fig:b} 
\begin{minipage}[t]{0.31\linewidth}
\centering
\includegraphics[width=1\linewidth]{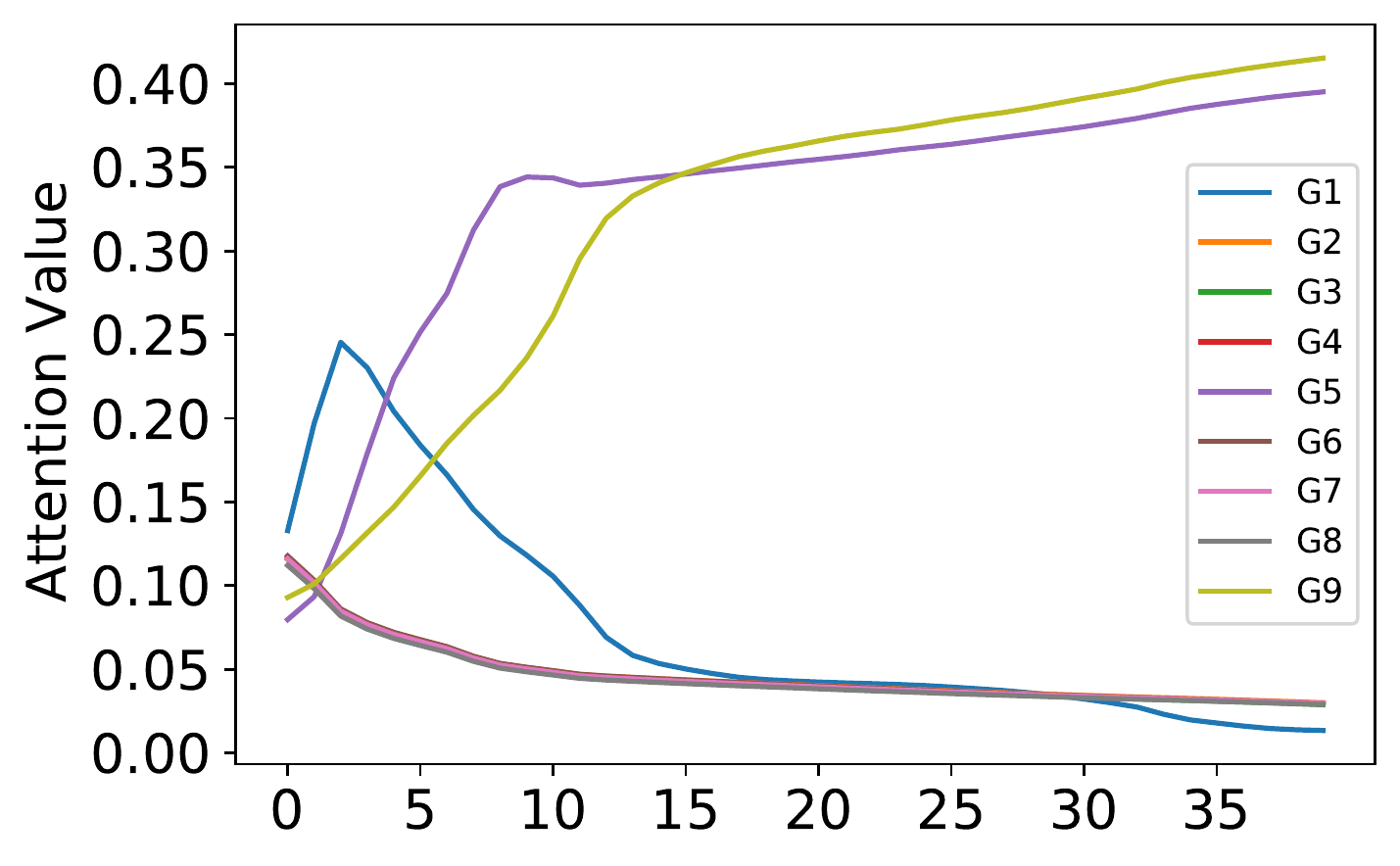}
\end{minipage}%
}%
\subfigure[ACM]{
\label{fig:c} 
\begin{minipage}[t]{0.32\linewidth}
\centering
\includegraphics[width=1\linewidth]{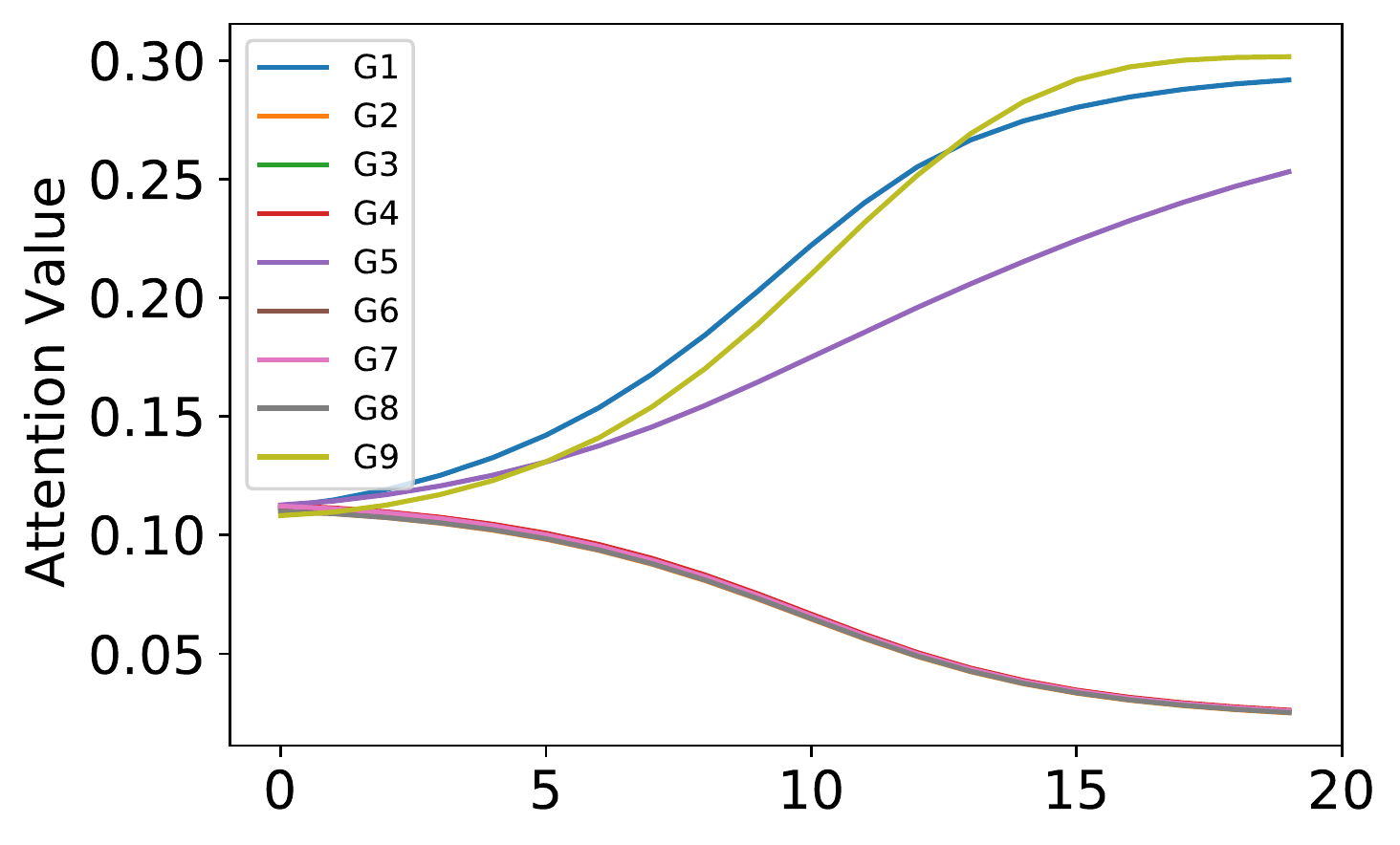}
\end{minipage}%
}%
\centering
\caption{The attention changing trends w.r.t epochs. (L/C=20)}
\label{attention_trends} 
\end{figure}

In Figure \ref{attention_trends}, we further analyze the changing trends of attention values for different input combinations in the increasing of training epochs. We report the results of ACM, UAI2010, and Citeseer datasets as examples, we can observe that the average attention values of different combinations gradually increase or decrease when training, and finally converge to a relatively stable value.
This phenomenon proves that the proposed attentional integrating model has a great fitting capability to learn attention values on different datasets.

\begin{figure}[htbp]
\centering
\subfigure[$G_1(A,X)$]{\includegraphics[width=4.0cm]{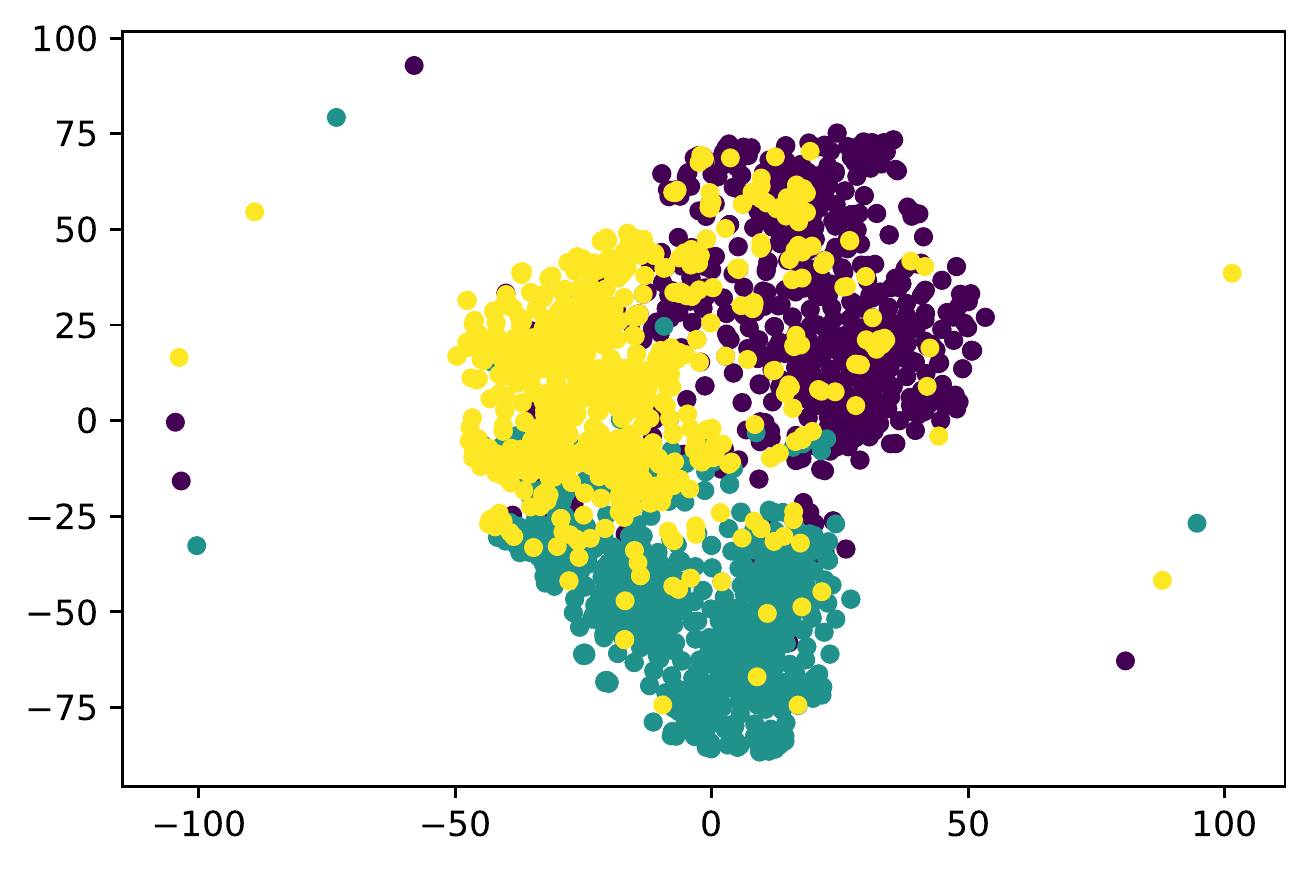}} 
\subfigure[$G_2(A,A_C)$]{\includegraphics[width=4.0cm]{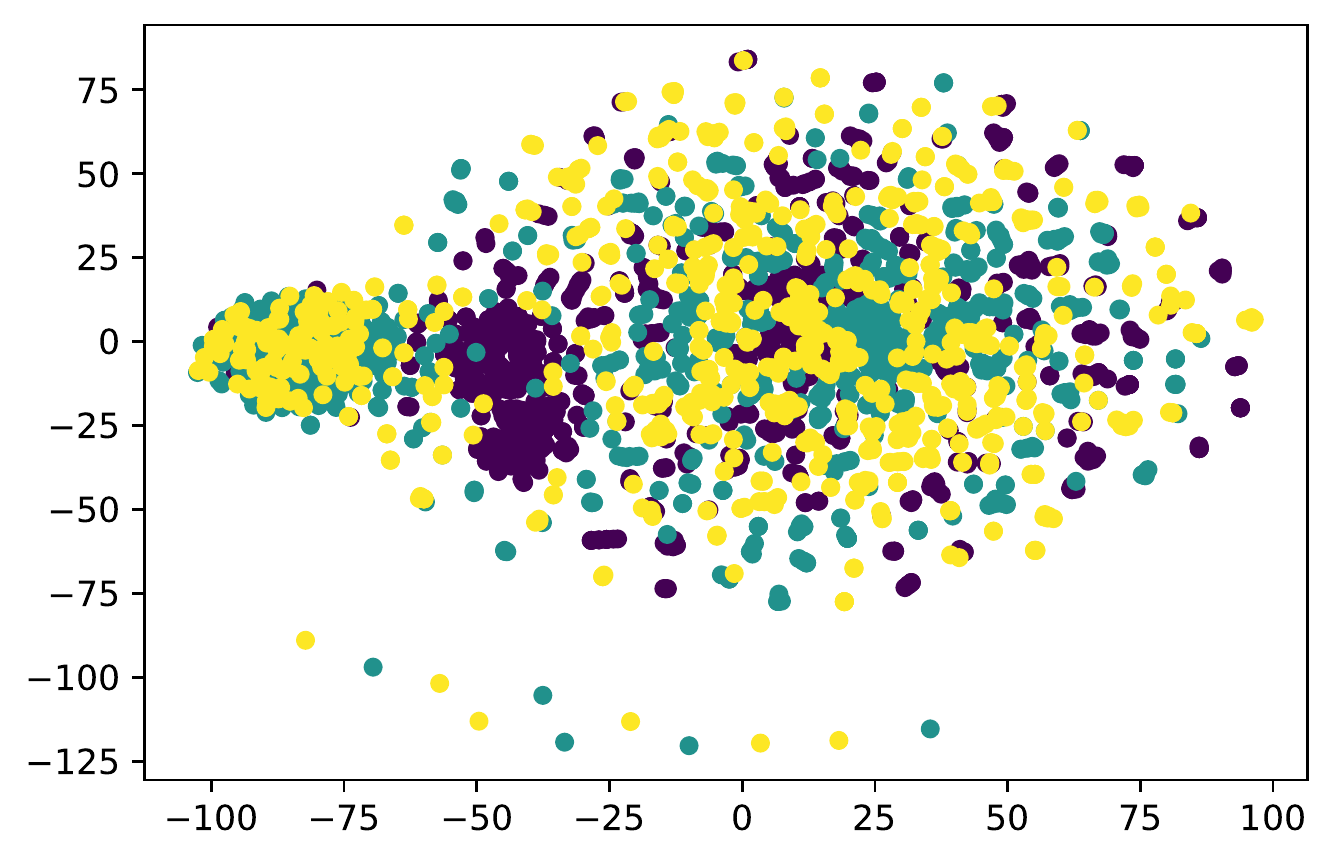}}
\subfigure[$G_3(A,X_C)$]{\includegraphics[width=4.0cm]{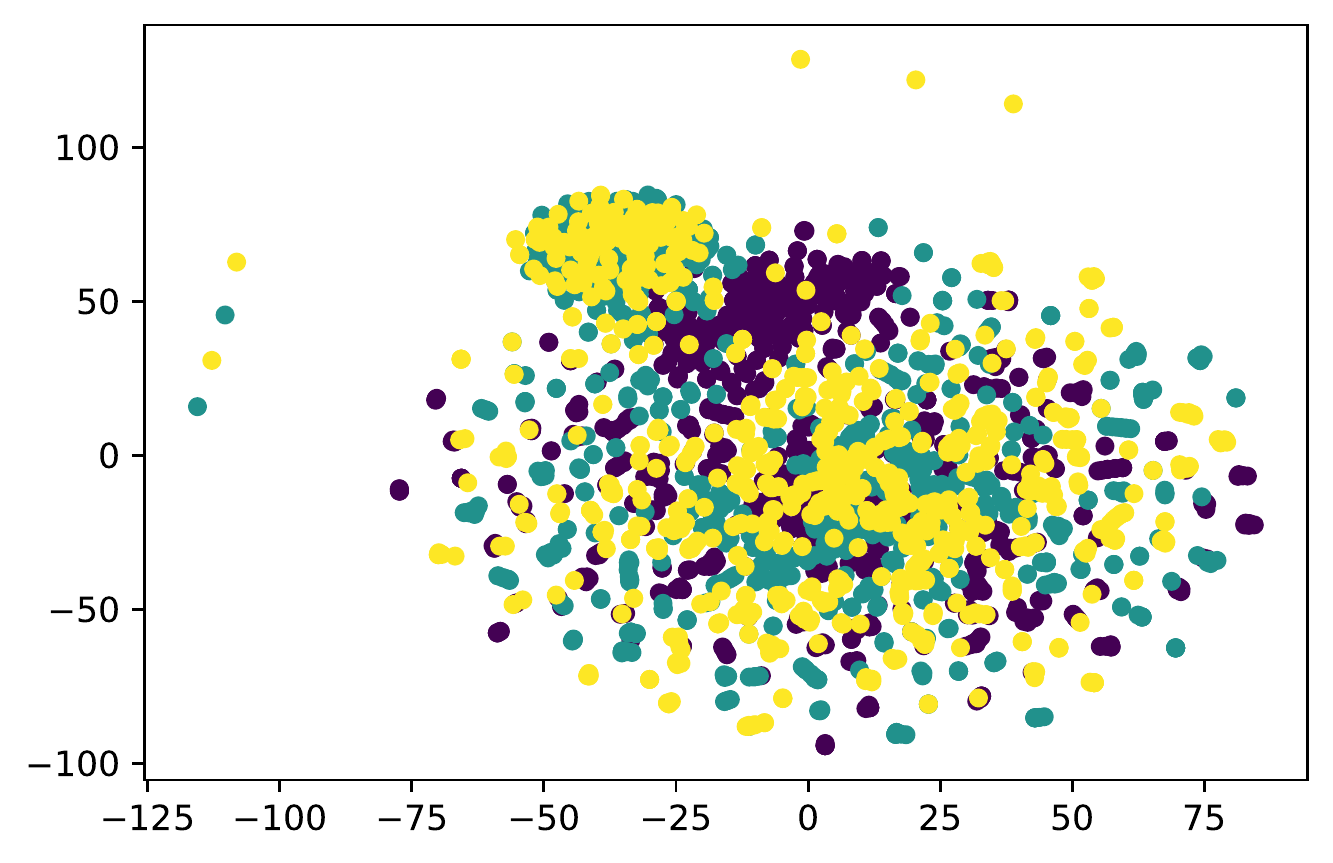}}
\\ 
\centering
\subfigure[$G_4(A_T,X)$]{\includegraphics[width=4.0cm]{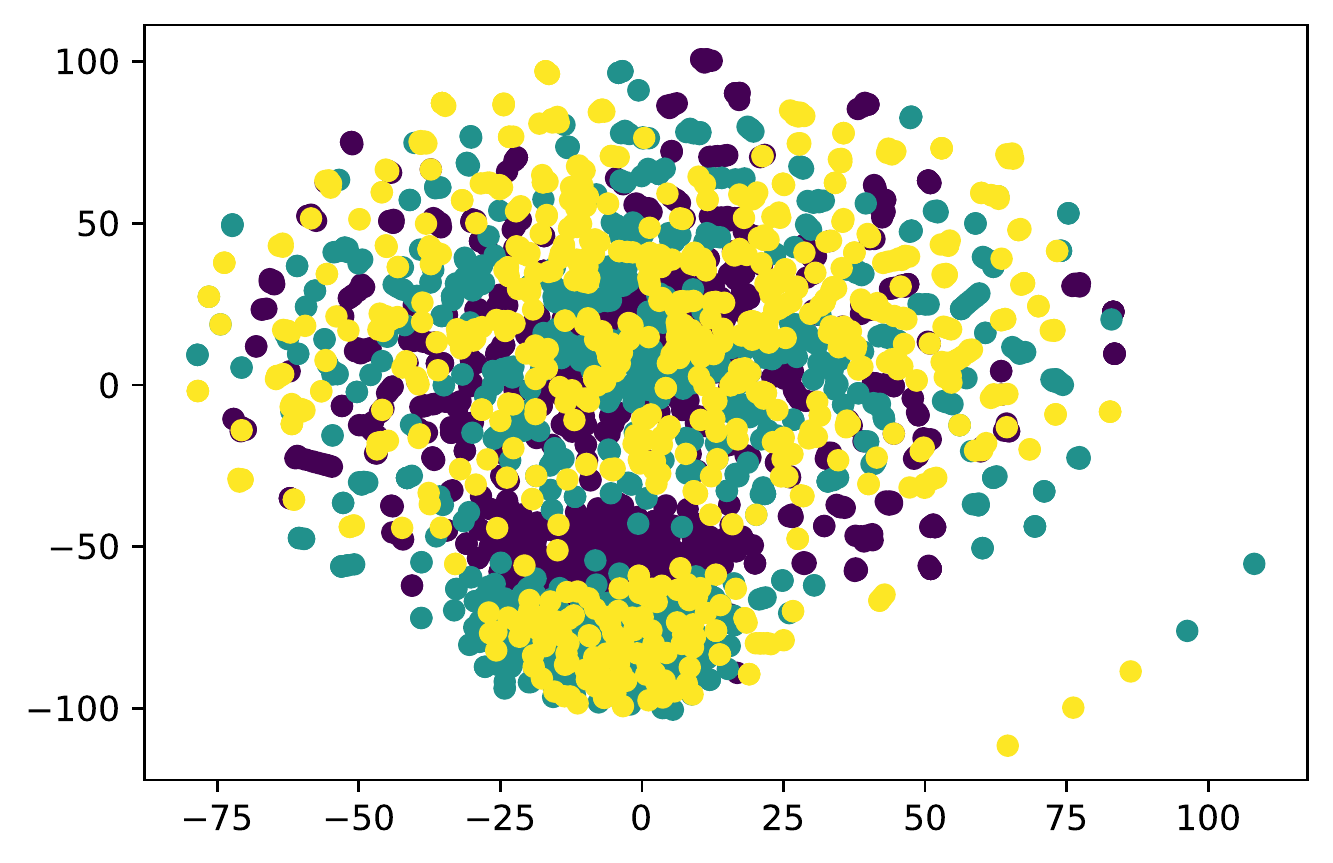}}
\subfigure[$G_5(A_T,A_C)$]{\includegraphics[width=4.0cm]{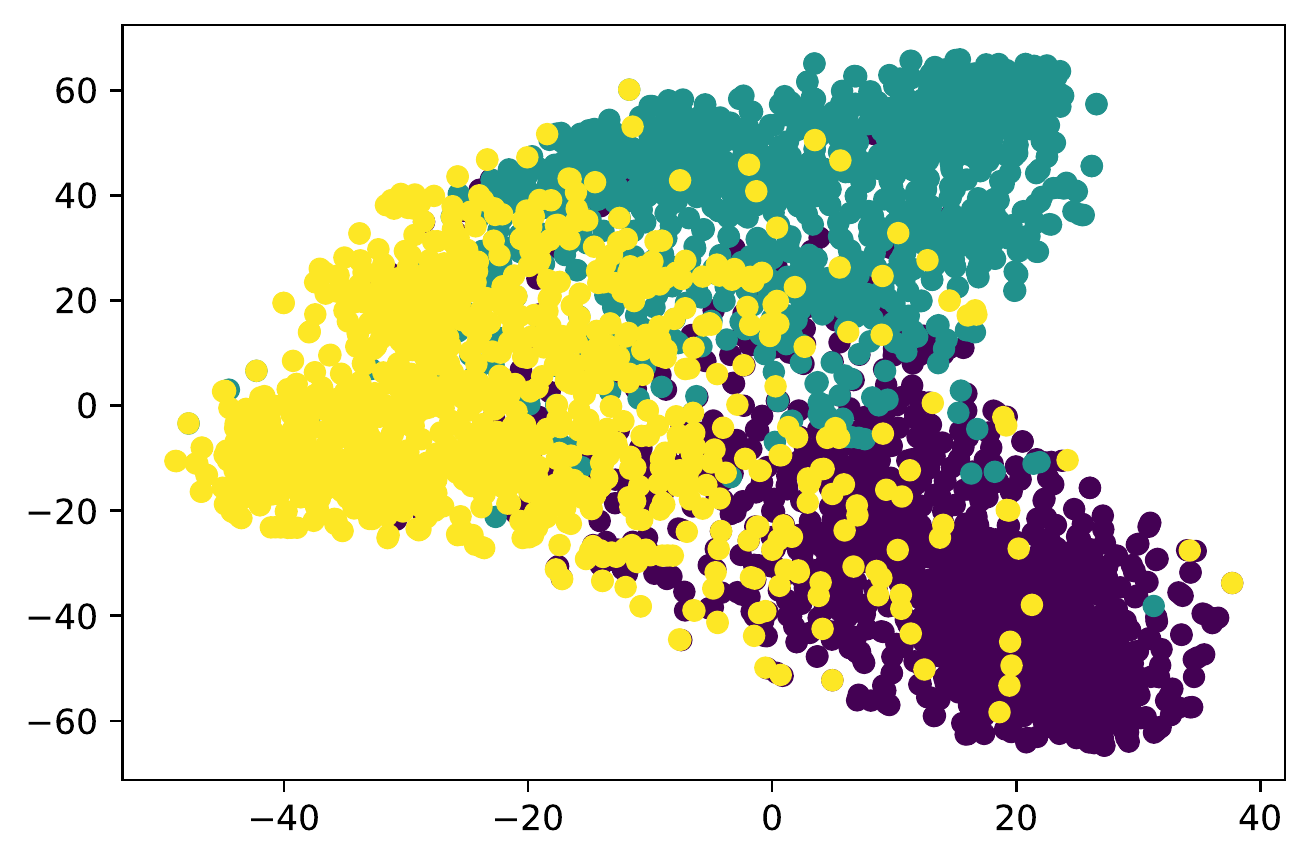}}
\subfigure[$G_6(A_T,X_C)$]{\includegraphics[width=4.0cm]{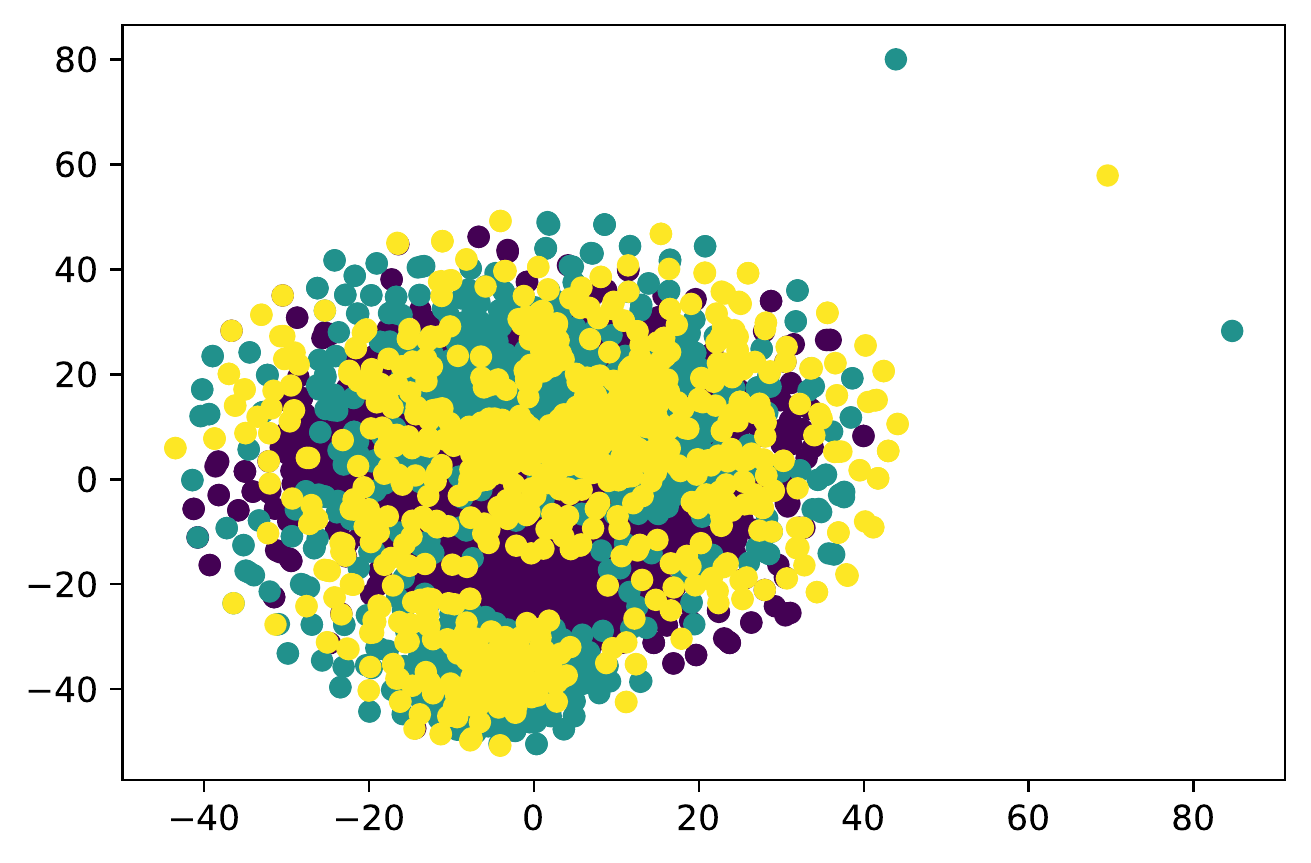}}
\\ 
\centering
\subfigure[$G_7(X_T,X)$]{\includegraphics[width=4.0cm]{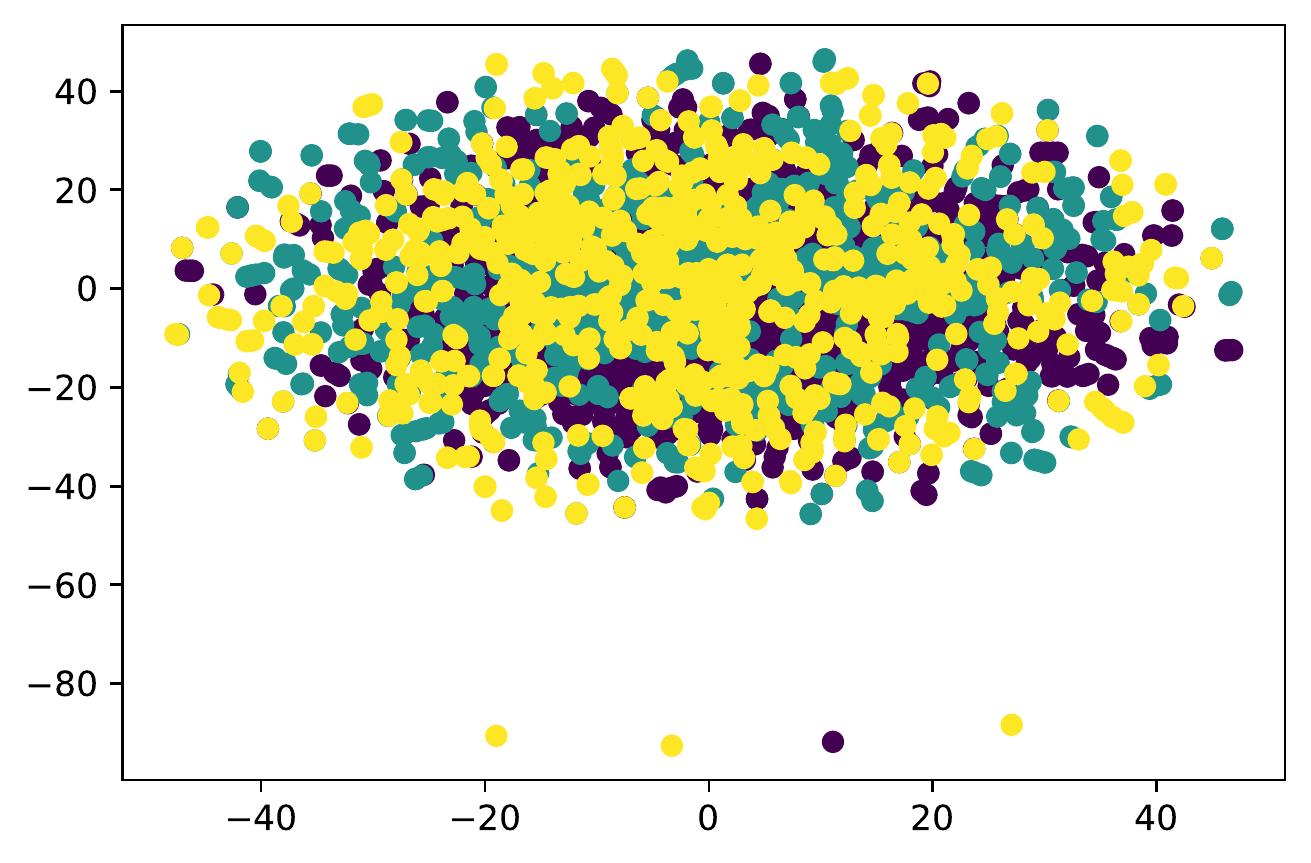}}
\subfigure[$G_8(X_T,A_C)$]{\includegraphics[width=4.0cm]{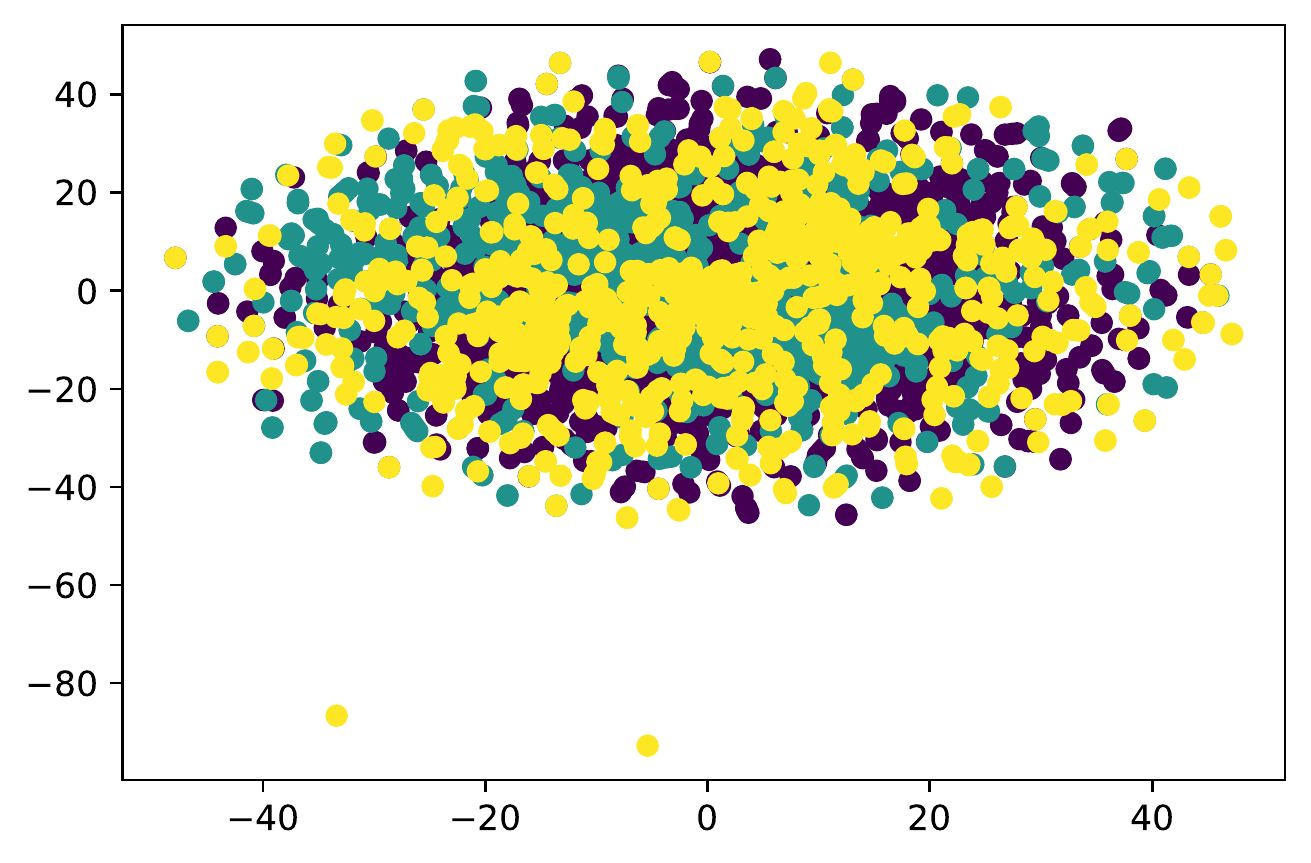}}
\subfigure[$G_9(X_T,X_C)$]{\includegraphics[width=4.0cm]{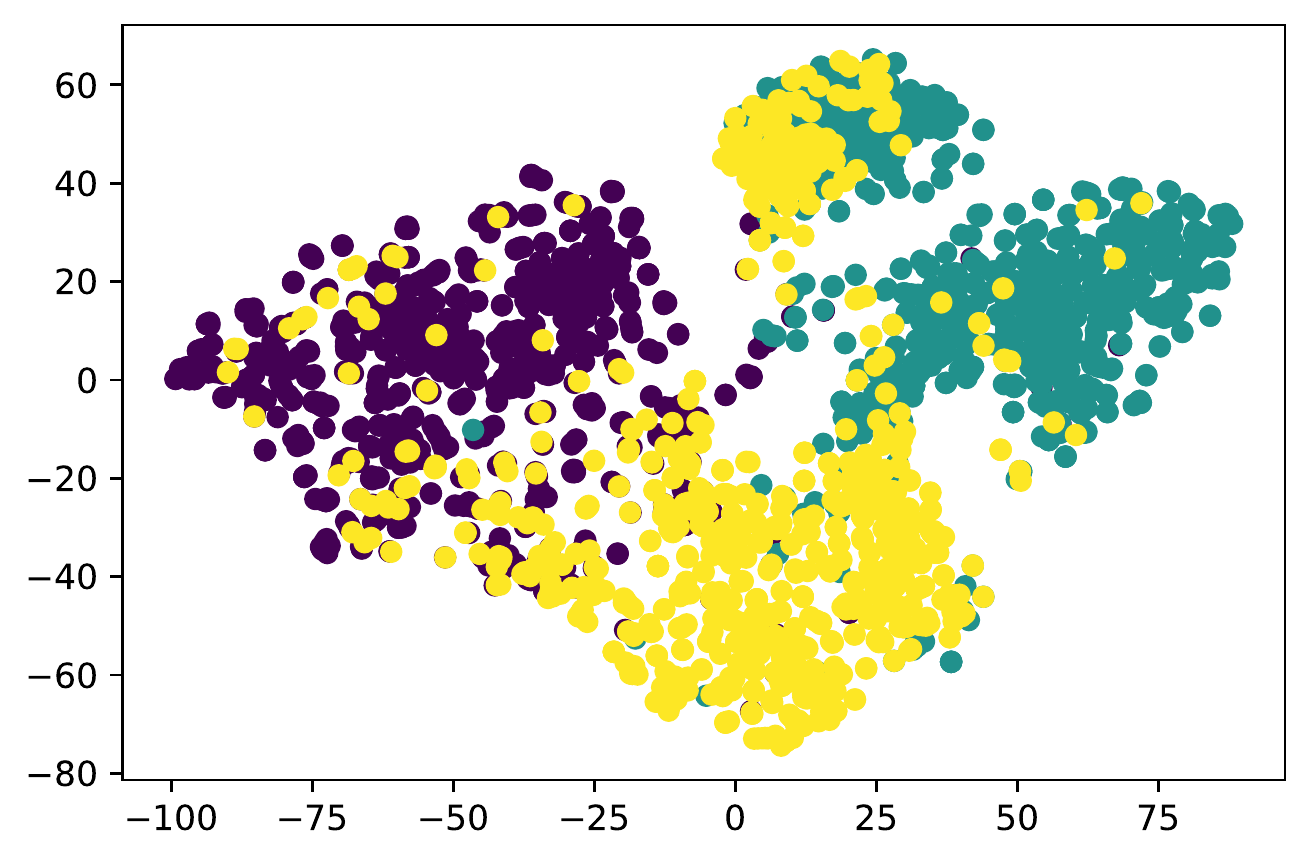}}

\caption{Visualization of hidden node embeddings on ACM datasets. (L/C=20)} 
\label{nine_embedded_visualization}
\end{figure}

We also demonstrate the distribution of the output node embeddings of nine combination inputs when the model has converged. Figure \ref{nine_embedded_visualization} shows the embedding distributions of the ACM dataset projected by t-SNE. It can be observed that the node embeddings $Z_1$, $Z_5$, and $Z_9$ encoded from $G_1(A,X)$, $G_5(A_T,A_C)$, and $G_9(X_T,X_C)$ is obviously well classified into three classes, so the learned attention of them in Figure \ref{attention_max} is larger than others. It proves that our designed graph features can also capture useful information for node classification and the attentional integration model can adaptively integrate different information from multiple input features to improve the final classification results.
Also, the distributions of nine node embeddings are significantly different from each other, showing the effectiveness of our designed disparity constraint in keeping the dependency of different embeddings.



\subsection{Parameter Sensitivity}

\begin{figure*}[htbp]
\centering
\subfigure[Citeseer]{
\label{fig:a} 
\begin{minipage}[t]{0.32\linewidth}
\centering
\includegraphics[width=1\linewidth]{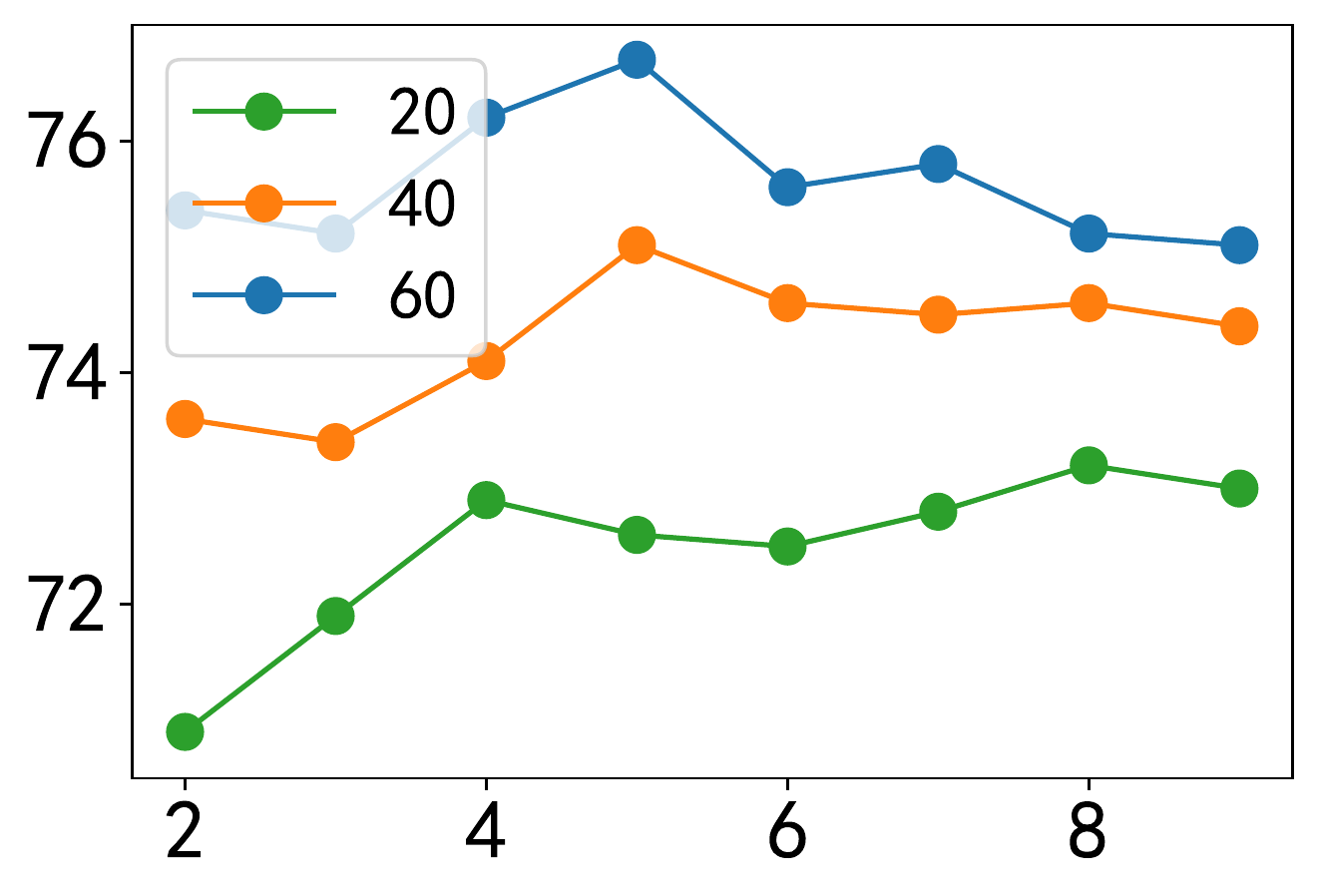}
\end{minipage}%
}%
\subfigure[UAI2010]{
\label{fig:a} 
\begin{minipage}[t]{0.32\linewidth}
\centering
\includegraphics[width=1\linewidth]{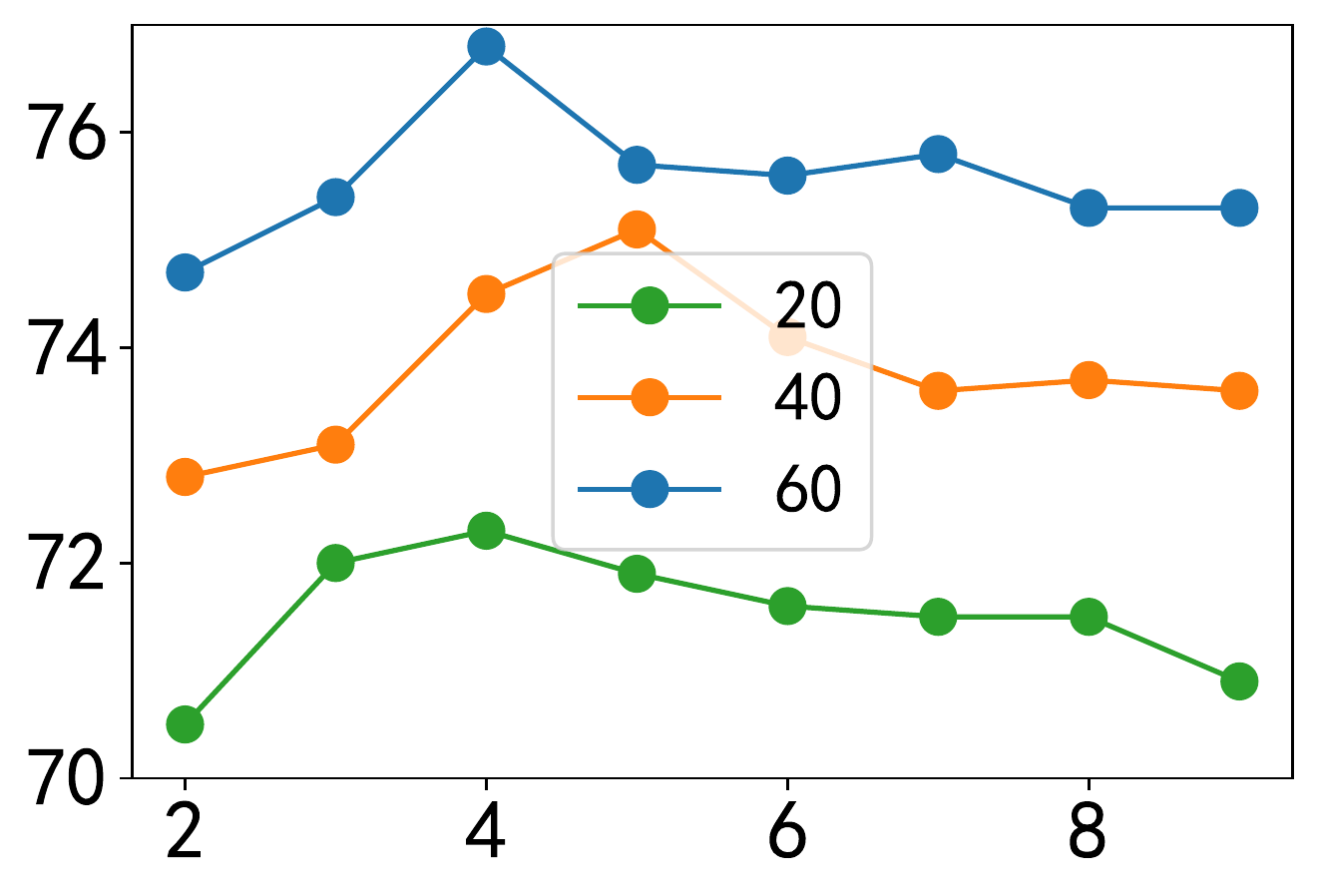}
\end{minipage}%
}%
\subfigure[ACM]{
\label{fig:b} 
\begin{minipage}[t]{0.32\linewidth}
\centering
\includegraphics[width=1\linewidth]{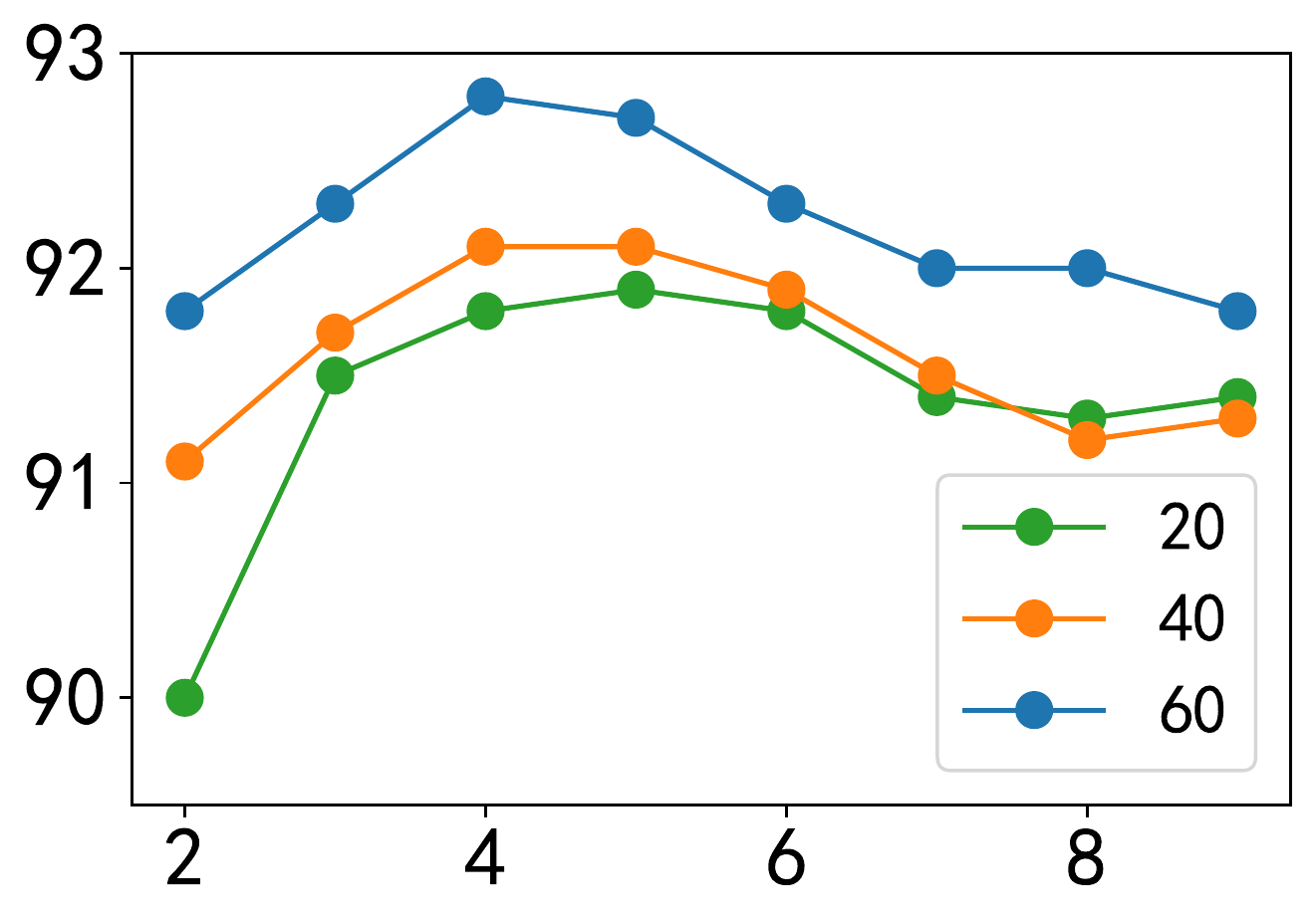}
\end{minipage}%
}%
\centering
\caption{Analysis of parameter k.}
\label{parameter_k} 
\end{figure*}

The parameter $k$ introduced in Section \ref{sec:DA} is used to adjust the sparsity of our augmented features $A_T$ and $X_T$.
In Figure \ref{parameter_k}, we evaluate how the $k$ impacts the performance of our method on ACM,  UAI2010,  and  Citeseer  datasets with the number of training nodes as 20/40/60, respectively. We report the ACC of our method with various numbers of $k$ ranging from 2 to 9 and other parameters remaining the same.
From the figures, we observe that when $k$ was small, the accuracy performance of our model is relatively limited, demonstrating that a smaller size of $k$ led to the augmented adjacency features sparser and information loss.
When $k$ is increased to 4 or 5, our model can gain the highest accuracy results. 
However, when $k$ is too large, the performance decreases slightly, which may probably because denser augmented adjacency features may introduce more noisy edges.
In summary, properly setting the size of $k$ can help to generate robust features to improve the performance of our method.

\section{Related Works}

Graph data augmentation has drawn increasing attention in graph learning recently, it can create new graph data to improve the generalization of graph models, especially the GNN models. 
Existing graph augmentations mainly focus on augmenting graph structures by modifying local graph structure\cite{rong2019dropedge,hamilton2017inductive,chen2018fastgcn}. \cite{zhou2020data} introduce data augmentation on graphs and present two heuristic algorithms: random mapping and motif-similarity mapping, to generate more weakly labeled data for small-scale benchmark datasets via heuristic modification of graph structures. \cite{kong2020flag} propose a simple but effective solution, FLAG, which iteratively augments node features with gradient-based adversarial perturbations during training, and boosts performance at test time. \cite{Wang2020AM} construct a feature
graph and propose an adaptive multi-channel graph convolutional networks to improve the node embeddings.
\cite{zhao2020data} shows that neural edge predictors can effectively encode class-homophilic structure to promote intra-class edges and demote inter-class edges in given graph structures, and their leverages these insights to improve performance in GNN-based node classification via edge prediction. \cite{wang2020nodeaug} present the Node-Parallel Augmentation scheme, that creates a ‘parallel universe’ for each node to conduct data augmentation.
\cite{spinelliefficient} proposed GINN that uses supervised and unsupervised data to construct a similarity map between points in the dataset, and rebuild them to expand the dataset.

\vspace{-0.2cm}

\section{Conclusion}
In this paper, we study to improve the performance of GCN on semi-supervised classification via graph data augmentation. We create new attribute and adjacency features base on original graph features and pairwise combine them as inputs for specific GCNs, then use attention mechanism and disparity constraint to integrate diverse information from the GCNs' outputs to the final node embeddings. From the experiments, our proposed method can better extract the rich information of graphs and improve the qualities of node representations.
\vspace{-0.2cm}

\section{Acknowledgments}
This work is supported in part by  the Natural Science Foundation of China under Grant No. 92046017, the Natural Science Foundation of China under Grant No. 61836013, Beijing Natural Science Foundation(4212030).

%
%
%
\renewcommand{\baselinestretch}{0.94} \normalsize
\bibliographystyle{splncs04}
\bibliography{bibfile.bib}

\end{document}